%% file: acl_latex.tex
\pdfoutput=1

\documentclass[11pt]{article}

\usepackage[dvipsnames,usenames,table]{xcolor}
\usepackage[]{acl}

\usepackage{times}
\usepackage{latexsym}

\usepackage{tabularx}

\usepackage[T1]{fontenc}

\usepackage[utf8]{inputenc}

\usepackage{microtype}

\usepackage{inconsolata}

\usepackage[textsize=tiny]{todonotes}
\usepackage{multicol}
\usepackage{graphicx}
\usepackage{enumitem}
\usepackage{xspace}
\usepackage{subfigure}
\usepackage{booktabs} %
\usepackage{multirow}
\usepackage{amsmath}
\usepackage{amsfonts}
\usepackage{xfp}
\usepackage{algorithm}
\usepackage{soul}

\usepackage{tabularx}
\usepackage{multirow}
\usepackage{enumitem} 
\usepackage{anyfontsize}
\usepackage{url}
\usepackage{subcaption}
\usepackage{xr}
\usepackage{tabu}
\usepackage{bbold}

\usepackage{MnSymbol,wasysym}
\usepackage{fontawesome}

\usepackage{times}
\usepackage{latexsym}
\usepackage{xspace}
\usepackage{wrapfig}

\usepackage{graphicx}
\usepackage{xfp}

\usepackage{array}
\usepackage{tcolorbox}
\usepackage{makecell}
\usepackage{soul}
\usepackage{stackengine}

\newtheorem{example}{Example}

\usepackage{array}
\newcolumntype{H}{>{\setbox0=\hbox\bgroup}c<{\egroup}@{}}

\input{macro.tex}

\addtocontents{toc}{\protect\setcounter{tocdepth}{-1}}

\title{Inferring from Logits: Exploring Best Practices for Decoding-Free Generative Candidate Selection}

\author{%
Mingyu Derek Ma$^{1}$\thanks{Equal contribution.}
\quad Yanna Ding$^{2}$$^*$
\quad Zijie Huang$^1$
\\ \textbf{Jianxi Gao}$^2$ 
\quad \textbf{Yizhou Sun}$^1$ 
\quad \textbf{Wei Wang}$^1$ \\
$^1$University of California, Los Angeles \quad $^2$Rensselaer Polytechnic Institute \\
\texttt{\{ma, zijiehuang, yzsun, weiwang\}@cs.ucla.edu}\\
\texttt{\{dingy6, gaoj8\}@rpi.edu}
}

\begin{document}

\maketitle

\begin{abstract}
    \input{content/0_abstract}
\end{abstract}

\input{content/1_intro}

\input{content/3_prelim}

\input{content/4_methods}
\input{content/5_setup}
\input{content/6_results}

\input{content/7_analysis}
\input{content/2_related_works}

\input{content/100_conclusion}

\input{content/200_acknowledgements}

\input{content/110_limitation}

\bibliography{neurips_2024,colm2024_conference,anthology,ma,ma_auto,custom,custom2,custom3,new_2406}

\clearpage
\appendix
\addtocontents{toc}{\protect\setcounter{tocdepth}{3}}

\hypersetup{linkcolor=black}

\renewcommand{\contentsname}{Appendix}

\tableofcontents %

\hypersetup{linkcolor=red}

\input{content/300_appendix}

\end{document}

%% file: macro.tex
\newcommand{\vtwo}[1]{#1}
\newcommand{\venlsp}[1]{}

\newcommand{\Skip}[1]{}

\newcommand{\ie}{\textit{i}.\textit{e}.\ }
\newcommand{\eg}{\textit{e}.\textit{g}.\ }

\newcommand{\secref}[1]{\S\ref{#1}}
\newcommand{\appref}[1]{Appendix~\ref{#1}}
\newcommand{\figref}[1]{Figure~\ref{#1}}
\newcommand{\tbref}[1]{Table~\ref{#1}}
\newcommand{\equationref}[1]{Equation~\ref{#1}}

\newcommand{\dotieconcat}[2]{%
  \text{\raisebox{.8ex}{$\smallfrown$}}%
}
\newcommand{\mypar}[1]{\noindent\textbf{#1}}

\newcommand{\spacemagic}[1]{}

\newcommand{\colorcellnull}[1]{\cellcolor{gray!20}---}

\newcommand{\colorcelldneg}[3]{\cellcolor{Mahogany! \fpeval{100 * (#3 - #1 - #2) / (50 - #2)}}-\fpeval{#3 - #1}}
\newcommand{\colorcelldpos}[3]{\cellcolor{Green! \fpeval{100 * (#1 - #3 - #2) / (50 - #2)}}+\fpeval{#1 - #3}}

\newcommand{\diffneg}[2]{
\colorcelldneg{#1}{0}{#2}
}
\newcommand{\diffpos}[2]{
\colorcelldpos{#1}{0}{#2}
}

%% file: content/0_abstract.tex
Generative Language Models rely on autoregressive decoding to produce the output sequence token by token. 
Many tasks such as preference optimization, require the model to produce task-level output consisting of multiple tokens directly by selecting candidates from a pool as predictions. 
Determining a task-level prediction from candidates using the ordinary token-level decoding mechanism is constrained by time-consuming decoding and interrupted gradients by discrete token selection. Existing works have been using decoding-free candidate selection methods to obtain candidate probability from initial output logits over vocabulary. Though these estimation methods are widely used, they are not systematically evaluated, especially on end tasks.
We introduce an evaluation of a comprehensive collection of decoding-free candidate selection approaches on a comprehensive set of tasks, including five multiple-choice QA tasks with a small candidate pool and four clinical decision tasks with a massive amount of candidates, some with 10k+ options. We evaluate the estimation methods paired with a wide spectrum of foundation LMs covering different architectures, sizes and training paradigms.
The results and insights from our analysis inform the future model design.

%% file: content/1_intro.tex
\section{Introduction}
\label{sec:intro}

Generative LMs respond to queries by generating tokens to form an output sequence and optimize themselves by learning to generate the correct tokens~\citep{Lewis2020BARTDenoisingSequencetoSequence}. The simplicity of token-level inference and optimization compromises its performance on end tasks, as there is a gap between the token-level paradigm and sequence-level task results and learning signals~\citep{Longpre2023FlanCollectionDesigninga,Ethayarajh2024KTOModelAlignment}.
Some tasks use generative LM to select the answer(s) from a given pool of options where each candidate answer is a natural language sequence. 
For example, 
multiple-choice QA considers answer options as the candidate pool~\citep{khashabi-etal-2020-unifiedqa}; the large collection of labels are candidate answers for extreme label classification tasks~\citep{amigo-delgado-2022-evaluating}; and disease ontology forms the candidate space for diagnosis tasks~\citep{Singhal2023LargeLanguageModels}.
The typical practice is to decode a complete output sequence and then match it with candidates~\citep{Mishra2022CrossTaskGeneralizationNatural}. However, selecting candidates using full decoding not only cuts off the gradient flow and disables direct optimization on decoded results but also limits the output bandwidth due to time-consuming discrete decoding.

\begin{figure}
    \centering
    \includegraphics[width=\linewidth]{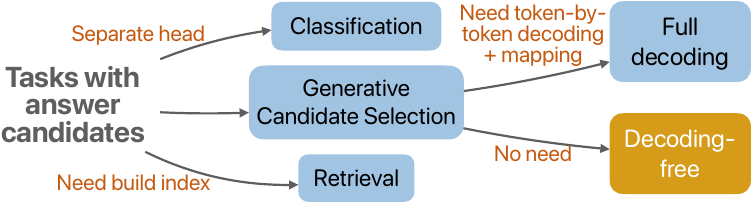}
    \caption{Decoding-free generative candidate selection produces results without training specialized parameters, building index, or token-by-token decoding.}
    \label{fig:teaser}
    \spacemagic{\vspace{-1.5em}}
\end{figure}

Existing works perform candidate selection without decoding for outcome-level optimization or efficient parallel predictions~\citep{Rafailov2023DirectPreferenceOptimization,ma2024memorize}. For example, \citet{Ma2023MitigatingBiasQuestion} calculate averaged logits of MCQA options to select an answer without decoding; 
\citet{Xu2023CanNLIProvidea} estimate the NLI result using logits of a single token.
Though these decoding-free candidate selection practices are widely used, there is no formal definition or clear investigation of the properties of each method. There is also no consensus about the guiding principles for deploying those methods under various tasks and data scenarios with diverse numbers, lengths, and complexity of candidate sequences. In this work, we formally define the decoding-free generative candidate selection task, and conduct the first systematic evaluation of typical practices on downstream tasks, reflecting the ultimate influence of the selection methods compared with conducting full decoding. 
Our systematic evaluation covers an \textbf{extensive collection of candidate selection methods}, including five decoding-free approaches to calculating candidate probability distribution from token logits, as well as encoding-only dense retrieval method and full decoding approach.
The effects of deploying various methods are evaluated with a \textbf{comprehensive set of downstream testbeds} widely used for LLM evaluation. The first type of testbed includes five multiple-choice QA tasks with broad target capabilities and candidate diversity, reflecting the candidate selection capabilities while the candidate pool is limited. We further increase the difficulty and examine the performance on tasks with massive numbers of candidates on expert-curated large ontology with 10k+ options for making diagnoses, procedure decisions, ordering lab tests, and prescribing medications.
We further dive into the characteristic shifts of candidate selection methods while using a \textbf{wide spectrum of foundational LMs}. The base models are diverse in terms of architectures (decoder-only or encoder-decoder), sizes (spanning from 137M to 11B), and training methods (pre-trained or instruction-tuned).

The evaluation provides insights into the properties of decoding-free candidate selection methods. The performance of the token-logits-based candidate representation is highly dependent on the properties of the pretrained LM, dataset domain difficulty, and candidate space diversity. Pure estimation methods can outperform non-instruction-tuned models due to the challenges faced by weak base models in handling certain question formats during decoding. In this case, estimation methods offer a more straightforward means of exhibiting knowledge through token logits. 
\vtwo{
The insights derived from our evaluation enable more informed and confident design for future estimation methods. We empirically demonstrate that the logits of the first output step are most informative; using selective tokens for estimations compromises the performance and scaling properties of various model sizes.
}

%% file: content/3_prelim.tex
\spacemagic{\vspace{-0.3em}}
\section{Problem Formulation}

\begin{figure*}[t]
  \centering \includegraphics[width=\textwidth]{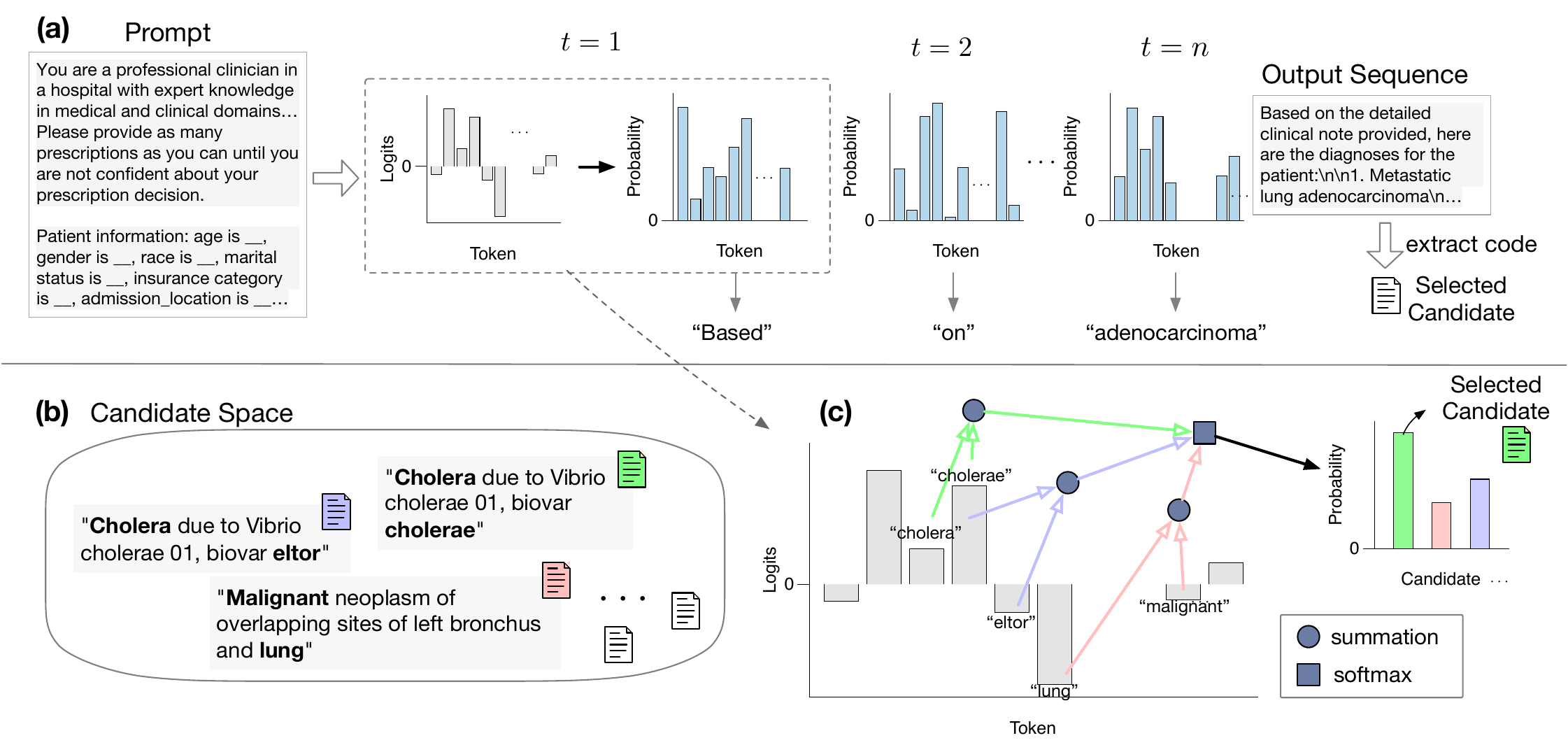}
  \spacemagic{\vspace{-1.7em}}
    \caption{ Demonstration of generative candidate selection methods. (a) Using full decoding for tasks with answer candidates by generating output sequences token-by-token. The task is to identify diagnoses given the patient's medical record. (b) Candidate space, consisting of coded diagnoses. (c) Using decoding-free generative candidate selection method for the diagnoses task based on prior-decoding logits over vocabulary. \label{fig:illustration} }
    \spacemagic{\vspace{-1.3em}}
\end{figure*}

\subsection{Decoding Paradigm of Generative LMs}
\label{sec:existing_paradigm}

The ordinary sequence-to-sequence formulation of generative LM takes the input sequence $seq_{in} = t_{1}^{in}, \ldots, t_{|seq_{in}|}^{in}$ and is expected to generate an output sequence $\hat{seq_{out}} = \hat{t_{1}^{out}}, \ldots, \hat{t_{|seq_{out}|}^{out}}$.
The output sequence generation involves encoding the input sequence to contextual vector representation \vtwo{(\ie, output of the final transformer block)}, and decoding the outputs following
\spacemagic{\vspace{-0.6em}}
\begin{equation}
\hat{seq_{out}} = f_{\operatorname{full-decode}}(f_{\operatorname{encode}}(seq_{in})).
\spacemagic{\vspace{-0.6em}}
\end{equation}
During inference, the decoding function $f_{\operatorname{full-decode}}$ involves \vtwo{$|seq_{out}|$} discrete decoding steps, \vtwo{in which each step produces one output token}. For $(k+1)-th$ step of decoding, which is conditioned on both the input sequence and $k$ generated tokens, the model produces logits $\mathbf{z}_k\in \mathbb{R}^{\lvert V \rvert}$ over the vocabulary $V$ \vtwo{after passing output of encoding through unembedding matrix}, and then obtain a probability distribution over the possible next token ($w \in V$) in the output sequence $P\left(w \mid t_{1:|seq_{in}|}^{in}, \hat{t}_{1:k}^{out}\right) = \operatorname{softmax}(\mathbf{z})$. Then a discrete token at this autoregressive decoding step is produced by \equationref{eq:decoding}.
\spacemagic{\vspace{-0.6em}}
\begin{equation}\label{eq:decoding}
\hat{t}_{k+1}^{out}=\operatorname{argmax}_{w \in V} P\left(w \mid t_{1:|seq_{in}|}^{in}, \hat{t}_{1:k}^{out}\right)
\end{equation}

\subsection{Candidate Selection with Answer Pools}
\label{sec:gen_tasks_candidate_pool}
When performing tasks using generative LMs, we include task instruction and query in the input sequence $seq_{in}$ and expect the derived answer of the query $\hat{ans}$ from the generated output sequence $\hat{seq_{out}}$ can match the ground-truth answer $ans$. Some tasks expect open-ended free generation \vtwo{where the final answer is the generated output} ($\hat{ans} = \hat{seq_{out}}$), such as translation, creative story generation and dialogue conversation~\citep{Wang2023Code4StructCodeGeneration,ma-etal-2023-parameter}. 
However, many tasks have an existing \vtwo{answer} candidate pool, and the output sequence needs to find a matched candidate as the final prediction. We notate the candidate pool as $C$ and a candidate as $c \in \{c_1, c_2, \ldots, c_{|C|}\}$ where $|C|$ is the total number of candidate options. Each candidate is a natural language sequence. The answer of the query has to be one of the candidate, \ie $ans \in C$. For example, answer options are candidates for multiple-choice question answering~\citep{talmor2018commonsenseqa}, segments of the input sentences are candidates of information extraction~\citep{sun2024learning,zhao2024dense,ma-etal-2024-star}, passages in large archives serve as candidates for information retrieval~\citep{Lewis2021RetrievalAugmentedGenerationKnowledgeIntensive}, and drugs within medication databases are candidates for prescription tasks~\citep{Yu2024LlaSMolAdvancingLarge,Ma2021HyperExpanTaxonomyExpansion}.

\subsection{Ordinary Approaches for Cand. Selection}
\label{sec:ordinary_approaches}

\textbf{Classification methods} train a $|C|$-way classification head with specialized parameters, where each candidate is treated as a class label. \textbf{Retrieval methods} first create an index with an encoded representation of each candidate. During prediction, the most matched candidates are retrieved, where the match is measured by the similarity between the candidate embedding and the query.

To select candidates using a \textbf{generative approach with full decoding}, the LM first generates a free-form output with discrete tokens $\hat{seq_{out}}$, then an additional mapping $f_{\operatorname{map}}$ from $\hat{seq_{out}}$ to candidates is needed to produce the probability over all candidates following \equationref{eq:full-decoding-answer}. \vtwo{This function can be a heuristic rule, semantic similarity matching, or manual processing.}
Then, the predicted answer $\hat{ans}$ where $\hat{ans} \in C$ is produced by $\hat{ans} = \operatorname{argmax}_{c \in C} P\left(c \mid t_{1:|seq_{in}|}^{in}\right)$.
\spacemagic{\vspace{-0.4em}}
\begin{equation}\label{eq:full-decoding-answer}
\resizebox{\columnwidth}{!}{$
\begin{split}
P\left(c \mid t_{1:|seq_{in}|}^{in}\right) &= \{p_{c^1}, p_{c^2}, ..., p_{c^{|C|}}\} \\
&= f_{\operatorname{map}}(f_{\operatorname{full-decode}}(f_{\operatorname{encode}}(seq_{in})))
\end{split}
$}
\spacemagic{\vspace{-0.3em}}
\end{equation}

\subsection{Decoding-free Generative Cand. Selection}

\vtwo{
At this point, we formally define the task of \textbf{decoding-free generative candidate selection}.
}
In \secref{sec:ordinary_approaches}, we introduce \textit{generative candidate selection using decoding}, where the answer is reflected by the output sequences. However, there are multiple severe limitations of decoding-based candidate selection. On the one hand, the discrete argmax operator for token selection interrupts the gradient flow, making applying objectives on task outcomes inefficient, such as using reinforcement learning with one-per-outcome sparse rewards instead of token-level feedback. On the other hand, the decoding process is time and resource-consuming, limiting the output bandwidth of generative LMs.

\textit{Decoding-free generative candidate selection} $f_{\operatorname{est}}$ is a function to produce the candidate prediction probability given $seq_{in}$ without discrete decoding. Given the encoded representation, the function calculates the logits of the first decoding step $\mathbf{z}_0$ (before the first discrete decoding) and then performs various approaches on top of the logits to estimate the probability of candidate outcomes. 
The assumption is that the model's intended preference over outcome candidates can be reflected in logits of tokens of the candidate sequences.

\begin{figure*}[t]
\resizebox{\textwidth}{!}{
\begin{tabular}{p{5cm}p{5.5cm}p{5.8cm}}
\begin{minipage}{5cm}
\spacemagic{\vspace{-0.7em}}
\begin{equation}\label{eq:kth_token}
p_{c^i} = \frac{\exp(logit(c_k^{i}))}{\sum_{j=0}^{|C|}\exp(logit(c_k^{j}))}
\spacemagic{\vspace{-0.5em}}
\end{equation}
\end{minipage}
&
\begin{minipage}{5.5cm}
\spacemagic{\vspace{-0.7em}}
\begin{equation}\label{eq:average_token}
p_{c^i} = \frac{\exp(\langle logit(c^{i}) \rangle)}{\sum_{j=0}^{|C|} \exp(\langle logit(c^{j}) \rangle)}
\spacemagic{\vspace{-0.5em}}
\end{equation}
\end{minipage}
&
\begin{minipage}{5.8cm}
\spacemagic{\vspace{-0.7em}}
\begin{equation}\label{eq:sum_token}
p_{c^i} = \frac{\exp(\sum_{k=0}^{|c^i|} logit(c^{i}_k))}{\sum_{j=0}^{|C|} \exp(\sum_{k=0}^{|c^j|} logit(c^{j}_k))}
\spacemagic{\vspace{-0.5em}}
\end{equation}
\end{minipage}
\end{tabular}
}
\spacemagic{\vspace{-0.8em}}
\end{figure*}

The estimation method directly produces the probability distribution over all candidate outputs following \equationref{eq:estimation-answer}. The predicted answer can be yielded by $\hat{ans} = \operatorname{argmax}_{c \in C} P\left(c \mid t_{1:|seq_{in}|}^{in}\right)$.
\spacemagic{\vspace{-0.6em}}
\begin{equation}\label{eq:estimation-answer}
P\left(c \mid t_{1:|seq_{in}|}^{in}\right) = f_{\operatorname{est}}(f_{\operatorname{encode}}(seq_{in}))
\spacemagic{\vspace{-0.6em}}
\end{equation}
Compared with \equationref{eq:full-decoding-answer}, the full decoding process $f_{\operatorname{full-decode}}$ (including probability calculation, argmax for token selection, decoding for the next token) and the mapping $f_{\operatorname{map}}$ are not conducted. 

We elaborate on the advantages and how it differ from ordinary approaches in \appref{sec:advantages_properties}.

%% file: content/4_methods.tex
\spacemagic{\vspace{-0.3em}}
\section{Generative Candidate Selection Methods}
\spacemagic{\vspace{-0.2em}}

We formally define candidate selection methods to be analyzed. We introduce ordinary approaches in \secref{sec:ordinary_methods}. For decoding-free generative candidate selection methods, there are three key design choices: 1) output step used to obtain the logits, \ie source of the raw logits data; 2) candidate sequence keyword selection, \ie what tokens represent a candidate option; 3) estimation methods $f_{\operatorname{est}}$ that produces probability over candidates given the candidate sequences and the logits at a certain output step. 
We discuss the first and second factors in \secref{sec:ablation_output_steps} and \secref{sec:candidate_keywords}, and define the estimation methods in \secref{sec:est_methods}.
\vtwo{
Though some of the estimation designs have been used by existing works, there is no justification or empirical analysis to support their design choices. 
To the best of our knowledge, this work is the first to provide a formal summary of these approaches and systematically investigate the properties of each design of generative candidate selection methods.
}

\subsection{Estimating Candidates Probabilities from Logits}
\label{sec:est_methods}

\mypar{Logits of $k$-th token.}
Each candidate is represented by a sequence with $|c^i|$ tokens, \ie, $c^i = c^i_1, \ldots,  c^i_{|c^i|}$.
From the logits across all tokens in the vocabulary $\mathbf{z}_0$, we calculate the logit for a single token (\eg, the first or the last token) of each candidate sequence and apply softmax to these selected logits to determine the probability $p_{c^i}$ of predicting a candidate $c^i$ among all candidates $C$ following \equationref{eq:kth_token}.
We consider two variants in our evaluation: \textit{first token} and \textit{last token estimation}.

\mypar{Averaged token logits.}
We average the logits across all tokens for each candidate and apply softmax to these averaged logits across all candidates to compute choice probabilities following \equationref{eq:average_token} where $\langle \cdot \rangle$ represents the averaging operator. 

\mypar{Sum of token logits.}
For each choice, we sum the logits across all tokens of the candidate sequence. We then apply softmax to these summed logits to determine the probability of selecting each choice following \equationref{eq:sum_token}.

\subsection{Ordinary Candidate Selection Methods}
\label{sec:ordinary_methods}

\mypar{Full decoding.}  
This method performs full decoding to obtain an output sequence following \secref{sec:existing_paradigm}, then uses a mapping function to find the corresponding predicted answer from the given candidate pool following \secref{sec:ordinary_approaches}. We use the task-specific mapping function introduced along with each dataset. Typical practices include using regular expressions to match patterns, such as ``Answers: '', and predicting the candidates with the highest semantic similarity with the output sequence.

\mypar{Dense retrieval.} We can also formulate the candidate selection task as a retrieval task and use dense passage retrieval as one of the reference models.
Specifically, the question and each candidate choice are embedded into a high-dimensional vector space using these encoders, and cosine similarity between them is computed.
This similarity score quantifies the relevance of each choice to the posed question and determines the probability of each choice being the correct answer.
For our experimental setup, we use the Facebook DPR question encoder and context encoder~\citep{karpukhin-etal-2020-dense} to generate embeddings of the questions and candidate choices.

%% file: content/5_setup.tex
\spacemagic{\vspace{-1em}}
\section{Evaluation Settings}
\label{sec:method} 
\spacemagic{\vspace{-0.3em}}

We apply the introduced candidate selection methods to ultimate downstream tasks to reflect their influence on end tasks. We introduce the selected tasks in \secref{sec:tasks} and base generative LMs in \secref{sec:base_gen_lms}.

\subsection{Testbed Tasks}
\label{sec:tasks}

We evaluate generative candidate selection methods on two typical types of candidate selection tasks. The first type contains a limited number of answer candidates so that all plausible choices can fit in the input prompt of the model if needed. The second type of task has a massive candidate pool with a large amount of candidates, which cannot fit in the input prompt. \vtwo{We show a comparison between these two settings in \tbref{table:eval_tasks_diff}.}
\vtwo{
We cover more details of the tasks in \appref{sec:testbed_details}, task statistics and properties in \appref{sec:testbed_stats} and the distribution of candidate sequences lengths in \appref{sec:candidate_len_distribution}.
}

\subsubsection{Tasks with Limited \# of Candidates}

We use five tasks with the provided candidate pools:  (1) CommonsenseQA~\citep{talmor2018commonsenseqa}, (2) MMLU~\citep{hendryckstest2021,hendrycks2021ethics}, (3) GPQA~\citep{rein2023gpqa}, (4) BIG-Bench~\citep{srivastava2022beyond}, and (5) ARC~\citep{allenai:arc}, covering commonsense questions, science and liberal arts subjects in different education levels, logical reasoning questions, etc. Instances in all datasets contain one correct option and multiple distractors. They vary in difficulty and number of candidates per instance (3 to 5 candidates).

\vtwo{
To require the model to answer in a specific format without intermediate thinking processes, we add specific instructions in the input prompt, as shown in \appref{sec:prompt_examples}.
When incorporating the candidate information, we use candidate sequences without indication heads (\eg A, B) to estimate the selection for decoding-free methods. For the full decoding baseline, candidate sequences with indicators are included in the input for a fair comparison.
For the mapping function $f_{\operatorname{map}}$ used by the full decoding, which converts output sequence $seq_{out}$ to candidate selection $\hat{ans}$, we capture the first occurrence of a candidate sequence or indication head with regular expressions as the prediction as further elaborated in \appref{sec:map_function_details}.
}

\subsubsection{Tasks with Massive \# of Candidates}
\label{sec:task_type_1}

We adapt four professional decision-making tasks introduced by \citet{clibench} where the answer has to fall in a large-scale expert-defined coding system as the second category testbeds. The goal is to select multiple candidates from the pool as the predicted clinical decisions.
They include: \textbf{(6) Diagnosis decisions on ICD-10-CM coding system.} Given the patient records of a hospital admission and the history diagnoses of the patient, the task aims to produce a set of diagnoses, each has to choose from chapters in the International Classification of Diseases (10th revision) coding system with 94k+ options.
\textbf{(7) Procedure decisions on ICD-10-PCS coding system.} The task determines a set of actions to be implemented to intervene in the patient's health status given the patient record at admission time. Candidates for procedures are codes in ICD-10-Procedure Coding System ontology with 85k+ options.
\textbf{(8) Lab orders on LOINC coding system.} Given the admission patient record, the task selects a set of lab items from the candidate pool of 3rd-level codes of the Logical Observation Identifiers Names and Codes system.
\textbf{(9) Prescriptions on ATC coding system.} The goal is to identify a set of medications, each coded as a pharmacological subgroup in the Anatomical Therapeutic Chemical classification system, to be prescribed to the patient given admission medical record.

\input{table/results_limited_candidates}

\spacemagic{\vspace{-0.3em}}
\subsection{Base Generative LMs}
\label{sec:base_gen_lms}
\spacemagic{\vspace{-0.2em}}
We assess decoding-free candidate selection approaches while using various pretrained generative language models, including both decoder-only models in the Mistral and LLaMA families, as well as encoder-decoder models in the Flan-T5 family.
For LLaMA~\citep{llama3modelcard} and Mistral~\citep{jiang2023mistral} models, we use both models without instruction tuning (\textit{LLaMA3 8B} and \textit{Mistral v0.3 7B}) and after instruction tuning (\textit{LLaMA3 Instruct 8B} and \textit{Mistral Instruct v0.3 7B}). Among Flan-T5 models, we use the 11B variant~\citep{https://doi.org/10.48550/arxiv.2210.11416}.
When preparing the input sequence $seq_{in}$, we apply the chat template for the models trained with the prompt template, and we append the generation prompt to indicate the start of the answer segment.

%% file: table/results_limited_candidates.tex
\newcommand{\refmodel}[1]{
#1\cellcolor{gray!20}
}
\begin{table}[t]
\centering  
\small
\resizebox{\linewidth}{!}{
\begin{tabular}{lr|r|r|r|r|r}
\toprule
\setlength\tabcolsep{3.95pt}
{\shortstack[l]{Model\\ (\# Param)}} & Method 
&  {{\shortstack[l]{Common-\\senseQA}}} & {MMLU}
& {GPQA}
&  {{\shortstack[l]{BIG-\\Bench}}} & {ARC} 
\\  
\midrule
 \multirow{5}{*}{{\shortstack[l]{LLaMA3\\(8B)}}} 
 & \refmodel{Decoding} & 
31.83 & 
36.53 &  
27.90 &  
34.00 &  
55.51  
\\
& First     & 
\diffpos{40.94}{31.83} & 
\diffneg{29.73}{36.53} & 
\diffneg{24.78}{27.90} & 
\diffneg{32.00}{34.00} & 
\diffneg{42.88}{55.51}   
\\ 
&Last & 
\diffpos{41.06}{31.83} & 
\diffneg{28.64}{36.53} & 
\diffneg{25.89}{27.90} & 
\diffneg{32.00}{34.00} & 
\diffneg{42.84}{55.51}  \\
& Average   &
\diffpos{35.07}{31.83} & 
\diffneg{32.20}{36.53} & 
\diffneg{22.54}{27.90} &
\diffneg{34.00}{34.00} & 
\diffneg{51.99}{55.51}  
\\  
& Sum & 
\diffpos{36.64}{31.83} & 
\diffneg{32.78}{36.53} & 
\diffneg{22.32}{27.90} & 
\diffpos{34.80}{34.00} & 
\diffneg{46.63}{55.51}     \\     
\midrule
\multirow{5}{*}{{\shortstack[l]{LLaMA3\\Instruct\\(8B)}}}  
& \refmodel{Decoding} & 
70.70 & 
58.86 &
27.68 & 
51.20 & 
91.70  \\
& First     & 
\diffneg{32.36}{70.70} & 
\diffneg{27.03}{58.86} & 
\diffneg{25.00}{27.68} & 
\diffneg{32.00}{51.20} & 
\diffneg{37.62}{91.70}  \\   
& Last   &
\diffneg{32.54}{70.70} & 
\diffneg{26.08}{58.86} & 
\diffneg{24.55}{27.68} & 
\diffneg{32.00}{51.20} & 
\diffneg{34.98}{91.70}  \\  
& Average &
\diffneg{34.62}{70.70} & 
\diffneg{26.31}{58.86} & 
\diffneg{20.54}{27.68} & 
\diffneg{35.60}{51.20} & 
\diffneg{38.96}{91.70}  \\ 
& Sum &
\diffneg{33.88}{70.70} & 
\diffneg{26.16}{58.86} & 
\diffneg{21.88}{27.68} & 
\diffneg{35.20}{51.20} & 
\diffneg{38.69}{91.70} \\
\midrule
\multirow{5}{*}{{\shortstack[l]{Mistral\\v0.3\\(7.3B)}}} 
 &\refmodel{Decoding}  & 
21.63 & 
25.51 &  
30.13 & 
30.00 & 
29.27   
\\
&First & 
\diffpos{48.42}{21.63} & 
\diffpos{30.17}{25.51} & 
\diffneg{26.34}{30.13} & 
\diffpos{32.40}{30.00} & 
\diffpos{48.68}{29.27}   \\ 
& Last & 
\diffpos{48.88}{21.63} & 
\diffpos{29.53}{25.51} & 
\diffneg{26.56}{30.13} & 
\diffpos{32.00}{30.00} & 
\diffpos{47.52}{29.27} \\ 
&Average & 
\diffpos{42.52}{21.63} & 
\diffpos{32.55}{25.51} & 
\diffneg{26.12}{30.13}  &
\diffpos{32.00}{30.00} & 
\diffpos{54.44}{29.27}  \\ 
&Sum & 
\diffpos{46.79}{21.63} & 
\diffpos{33.13}{25.51} & 
\diffneg{24.55}{30.13} &
\diffpos{31.20}{30.00} &
\diffpos{53.32}{29.27}  \\
\midrule
\multirow{5}{*}{{\shortstack[l]{Mistral\\Instruct\\v0.3\\(7.3B)}}}
&\refmodel{Decoding}&
65.12 &  
52.06 &   
29.02 &  
47.60 &   
86.43 \\
&First & 
\diffneg{46.78}{65.12} & 
\diffneg{30.54}{52.06} & 
\diffneg{25.00}{29.02} & 
\diffneg{31.60}{47.60} & 
\diffneg{52.30}{86.43}   \\ 
& Last & 
\diffneg{46.60}{65.12} &
\diffneg{29.30}{52.06} &
\diffneg{26.56}{29.02} &
\diffneg{32.00}{47.60} &
\diffneg{50.51}{86.43}   \\ 
&Average & 
\diffneg{44.38}{65.12} & 
\diffneg{32.58}{52.06} & 
\diffneg{24.11}{29.02} & 
\diffneg{31.60}{47.60} & 
\diffneg{59.35}{86.43}   \\ 
&Sum & 
\diffneg{47.70}{65.12} & 
\diffneg{33.19}{52.06} & 
\diffneg{23.44}{29.02} & 
\diffneg{31.20}{47.60} & 
\diffneg{59.35}{86.43}   \\
\midrule
\multirow{5}{*}{{\shortstack[l]{Flan-T5\\XL (11B)}}}
&\refmodel{Decoding} &
97.48 &  
48.36 &  
25.45 &  
65.20 &  
89.25  \\
&First &  
\diffneg{52.52}{97.48} & 
\diffneg{26.87}{48.36} & 
\diffneg{24.11}{25.45} & 
\diffneg{32.40}{65.20} & 
\diffneg{46.81}{89.25}  \\ 
& Last &
\diffneg{51.95}{97.48} & 
\diffneg{27.11}{48.36} & 
\diffneg{24.55}{25.45} & 
\diffneg{32.00}{65.20} & 
\diffneg{44.76}{89.25}   \\ 
&Average &
\diffneg{57.43}{97.48} & 
\diffneg{29.82}{48.36} & 
\diffneg{23.66}{25.45} & 
\diffneg{34.00}{65.20} & 
\diffneg{55.42}{89.25}  \\ 
&Sum & 
\diffneg{50.64}{97.48} & 
\diffneg{26.89}{48.36} & 
\diffneg{21.43}{25.45} & 
\diffneg{33.20}{65.20} & 
\diffneg{49.35}{89.25}  \\
\midrule
{{\shortstack[l]{DPR}}} 
& \refmodel{Retrieval}
& 32.07 &  
27.15 & 
25.22 & 
30.80 & 
39.76    \\ 
\midrule
-& \refmodel{Random}
& 20.00 &
25.00 & 
25.00 & 
33.33 & 
25.00    
\\ \bottomrule
\end{tabular}} 
\caption{Accuracy for each method across five multiple-choice QA datasets with a limited number of candidates per question. For generative candidate selection methods without decoding, we report the performance gap compared with full decoding. Methods that underperform or outperform full decoding are highlighted with \colorbox{red!10}{red} or \colorbox{green!10}{green} background.
\label{tab:mcq}
}
\spacemagic{\vspace{-1.3em}}
\end{table}

%% file: content/6_results.tex
\spacemagic{\vspace{-1em}}
\section{Experiments Results
\label{sec:eval}}
\spacemagic{\vspace{-0.3em}}

\begin{figure*}[t]
  \centering \includegraphics[width=0.8\textwidth]{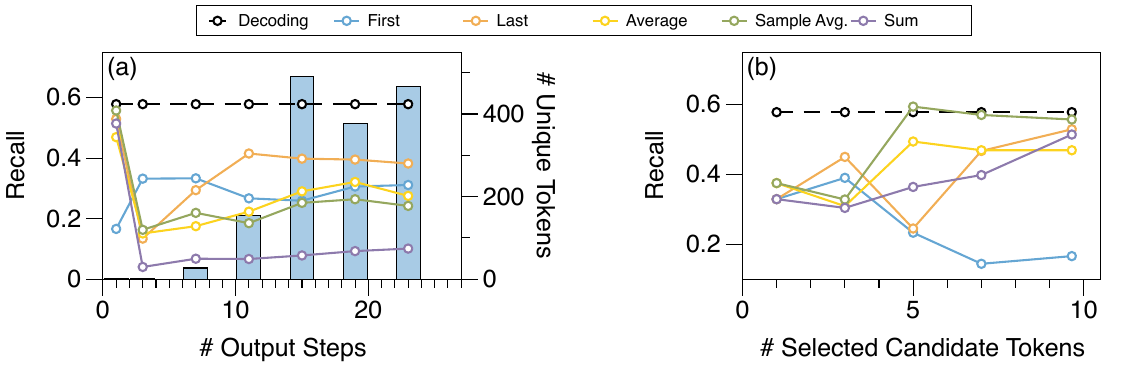}
  \spacemagic{\vspace{-0.7em}}
    \caption{\vtwo{Recall for the diagnosis task of various estimation methods using LLaMA3 8B Instruct while (a) using logits obtained from different output steps and (b) using logits calculated over different numbers of essential tokens selected from candidate sequences. The bars in (a) indicate the unique tokens for full decoded sequences at corresponding output steps, reflecting the diversity of the decoded tokens.}
    \label{fig:ablation_output-step_cand-seq} }
    \spacemagic{\vspace{-1em}}
\end{figure*}

\vtwo{
\tbref{tab:mcq} and \tbref{tab:clibench} show the candidate selection performance on tasks with limited and large-scale candidate pools, respectively. The runtime of each method is shown in \tbref{tab:runtime}.
Given longer candidate sequences, we introduce a new decoding-free candidate selection approach named \texttt{Sample Avg.}, which calculates average logits for \textit{every other} token in candidate sequences.
Besides the analysis for output steps, candidate token selection (\figref{fig:ablation_output-step_cand-seq}), candidate length and model sizes (\figref{fig:candidate-length_model-size}), we additionally demonstrate that adding chat template for instruction-tuned model hurts the estimation performance in \appref{sec:chat_template_study}, the effect of chain-of-thought reasoning in \appref{sec:cot_results}, additional ablation study on candidate length in \appref{sec:additional_candidate_length_ablation} and performance breakdown in \appref{sec:performance_breakdown}.
}

\input{table/results_massive_candidates}

\subsection{Characteristics of Generative Candidate Selection Methods}

\vtwo{
\mypar{Insight 1: Estimation methods provide reasonable initial guesses for challenging tasks and decision intuition especially when full decoding is weak.}
In \tbref{tab:mcq} with limited candidates, for more challenging datasets such as GPQA, decoding-free candidate selection approaches (also referred to as ``estimation methods'') provide a reasonable initial guess and do not necessarily perform significantly worse than full decoding.
Compared to full decoding, estimation methods even provide better performance for CommonsenseQA using LLaMA3 and all MCQA tasks except GPQA using Mistral v0.3. We observe these two models still struggle to handle the format for answering the question for some tasks during decoding, so it is hard to project its knowledge to interpretable results since the only surface to represent knowledge, outputting sequences, is not working for a weak base model. While knowledge by estimation methods is easier to exhibit through token logits.

For the results on clinical decisions with massive candidates presented in \tbref{tab:clibench}, all methods experience a decrease in performance on these more challenging tasks compared to the ones with a limited candidate space. Among decoder-only models, estimation methods can outperform full decoding for lab orders and prescriptions, particularly in non-instruction-tuned variants. Specifically, all estimation approaches surpass Mistral v0.3 in lab test orders, with \texttt{Sample Avg.} achieving the highest increase of 29.25 points compared to full decoding. 
Additionally, four of the estimation methods outperform LLaMA3 and Mistral v0.3's decoding methods in prescription decision making. 
The estimation methods provide hints of candidate selections in token logits. It is particularly useful when the full decoding approach of non-instruction-tuned models struggles to follow instructions (as shown in qualitative analysis in \appref{appendix:example}). When the model is able to understand the instruction and produce reasonable output (using instruction-tuned models), full decoding is still better than estimation. This aligns with our observation in \tbref{tab:mcq}.
To summarize, full decoding may impede the accurate selection of candidates, especially for non-instruction-tuned models, whereas decoding-free methods can provide an initial guess in some cases since they are not influenced by trajectory biases.
}

\mypar{Insight 2: Estimation methods lag behind when full decoding performs well.}
In \tbref{tab:mcq}, we observe an overall drop in performance when using estimation approaches, especially when the full decoding method achieves reasonable accuracy. This aligns with the intuition that estimation methods rely solely on the logits without capturing the token dependencies within the output.

\vtwo{
\mypar{Insight 3: Estimation results are similar before or after instruction tuning.}
Though instruction-tuned models tend to achieve better results than non-instruction-tuned ones with full decoding, the estimated selection results using the models of the same family do not have a large gap (LLaMA3 and Mistral compared with their instruct variants). This indicates that instruction tuning benefits the decoding method a lot while making no significant difference for decoding-free methods.
}

\mypar{Insight 4: Each candidate selection method excels under different conditions.}
\label{sec:results_type1}
The effectiveness of a candidate representation depends heavily on the specific LLM and dataset. For instance, when using the CommonsenseQA dataset, selecting by \texttt{Sum} logits is the best for Mistral Instruct v0.3, while the \texttt{Average} method performs best for Flan-T5. 
The difference in performance between the two single-token-based estimation methods (\texttt{First} and \texttt{Last}) is small, likely due to the limited length of most candidates.
The DPR model without fine-tuning performs similarly to random guessing on more difficult datasets such as GPQA and BIG-Bench as the retrieval model is designed for semantic similarity instead of reasoning. 
Both the capabilities of the pre-trained LM and the choice of representative tokens play crucial roles in accurate candidate selection.

\mypar{Insight 5: Decoding-free estimation is much more efficient than full decoding.}
As shown in \tbref{tab:runtime}, the decoding-free estimation methods bring significant speedups across all nine tasks, especially for tasks with large candidate pools (25.1x to 57.6x speedup).

\input{table/runtime}

%% file: table/results_massive_candidates.tex
\begin{table}[t]
    \centering  
    \small
    \resizebox{\linewidth}{!}{
        \begin{tabular}{lr|r|r|r|r }
            \toprule 
            {\shortstack[l]{Model\\ (\# Param)}} & Method  
            & \shortstack[l]{Diag-\\noses}  
            & \shortstack[l]{Proc-\\edures}   
            & \shortstack[l]{Lab \\ Orders}   
            & \shortstack[l]{Presc-\\riptions}   
            \\  \midrule
            \multirow{6}{*}{\shortstack[l]{LLaMA3\\ (8B)}} 
            & \refmodel{Decoding}
            & 34.86 
            & 9.42  
            & 36.52 
            & 31.00
            \\
            & First
            & \diffneg{15.69}{34.86}
            & \diffneg{7.97}{9.42} 
            & \diffpos{46.88}{36.52}
            & \diffpos{42.05}{31.00} \\
            & Last
            & \diffneg{11.14}{34.86} 
            & \diffneg{1.81}{9.42} 
            & \diffneg{25.36}{36.52}
            & \diffpos{34.81}{31.00}  \\
            & Average
            & \diffpos{37.03}{34.86} 
            & \diffneg{2.71}{9.42} 
            & \diffneg{34.66}{36.52}
            & \diffpos{40.61}{31.00}  \\
            & Sample Avg.
            & \diffneg{31.81}{34.86} 
            & \diffneg{2.26}{9.42} 
            & \diffpos{44.15}{36.52}
            & \diffpos{39.05}{31.00}  \\
            & Sum
            & \diffneg{17.80}{34.86} 
            & \diffneg{1.05}{9.42} 
            & \diffneg{32.42}{36.52}
            & \diffneg{24.92}{31.00}  \\
            \midrule
            \multirow{6}{*}{{\shortstack[l]{LLaMA3 \\Instruct \\(8B)}}}
            & \refmodel{Decoding} 
            & 57.82  
            & 27.04   
            & 47.04   
            & 49.74
            \\
            & First
            & \diffneg{16.60}{57.82} 
            & \diffneg{0.14}{27.04} 
            & \diffneg{35.49}{47.04}
            & \diffneg{42.35}{49.74}  \\
            & Last
            & \diffneg{52.93}{57.82} 
            & \diffneg{0.62}{27.04} 
            & \diffneg{20.81}{47.04}
            & \diffneg{33.98}{49.74}  \\
            & Average
            & \diffneg{46.95}{57.82} 
            & \diffneg{0.62}{27.04} 
            & \diffneg{29.45}{47.04}
            & \diffneg{42.06}{49.74}  \\
            & Sample Avg.
            & \diffneg{55.74}{57.82} 
            & \diffneg{1.05}{27.04} 
            & \diffneg{30.25}{47.04}
            & \diffneg{37.56}{49.74}  \\
            & Sum
            & \diffneg{51.37}{57.82} 
            & \diffneg{0.33}{27.04} 
            & \diffneg{31.09}{47.04}
            & \diffneg{24.22}{49.74}  \\
            \midrule
            \multirow{6}{*}{{\shortstack[l]{Mistral \\v0.3\\ (7.3B)}}}
            & \refmodel{Decoding} 
            & 22.83  
            & 3.33  
            & 18.17  
            & 12.83 
            \\
            & First
            & \diffneg{11.98}{22.83} 
            & \diffneg{2.51}{3.33} 
            & \diffpos{37.81}{18.17}
            & \diffpos{28.96}{12.83}  \\
            & Last
            & \diffpos{24.90}{22.83}  
            & \diffneg{0.00}{3.33} 
            & \diffpos{44.47}{18.17}
            & \diffpos{25.72}{12.83}  \\
            & Average
            & \diffneg{12.57}{22.83}  
            & \diffneg{0.85}{3.33} 
            & \diffpos{46.56}{18.17}
            & \diffneg{1.39}{12.83} \\
            & Sample Avg.
            & \diffneg{16.90}{22.83}  
            & \diffneg{1.45}{3.33} 
            & \diffpos{47.42}{18.17}
            & \diffpos{28.98}{12.83}  \\
            & Sum
            & \diffneg{15.37}{22.83} 
            & \diffneg{0.75}{3.33}
            & \diffpos{46.55}{18.17}
            & \diffpos{21.16}{12.83} \\
            \midrule
            \multirow{6}{*}{{\shortstack[l]{Mistral \\Instruct \\v0.3 \\(7.3B)}}}
            &\refmodel{Decoding}
            & 63.81 
            & 24.99   
            & 43.96  
            & 40.53  
            \\
            & First
            & \diffneg{16.12}{63.81}  
            & \diffneg{2.39}{24.99}  
            & \diffpos{45.37}{43.96} 
            & \diffneg{12.70}{40.53}    \\
            & Last
            & \diffneg{26.38}{63.81}  
            & \diffneg{0.00}{24.99}  
            & \diffneg{20.77}{43.96}  
            & \diffneg{21.83}{40.53}  \\
            & Average
            & \diffneg{39.72}{63.81}  
            & \diffneg{1.62}{24.99} 
            & \diffneg{29.83}{43.96}  
            & \diffneg{29.09}{40.53}   \\
            & Sample Avg.
            & \diffneg{46.90}{63.81} 
            & \diffneg{2.20}{24.99}  
            & \diffpos{54.32}{43.96} 
            & \diffneg{40.24}{40.53}  \\
            & Sum
            & \diffneg{28.59}{63.81}  
            & \diffneg{0.10}{24.99}  
            & \diffneg{30.52}{43.96}  
            & \diffneg{40.25}{40.53}   \\
            \midrule
            \multirow{6}{*}{{\shortstack[l]{Flan-T5\\XL (11B)}}}
            &\refmodel{Decoding} 
            & 10.22  
            & 0.73  
            & 8.43  
            & 4.41 \\
            & First
            & \diffpos{25.99}{10.22}  
            & \diffpos{2.91}{0.73}   
            & \diffpos{49.45}{8.43}  
            & \diffpos{36.55}{4.41}   \\
            & Last
            & \diffpos{30.02}{10.22}  
            & \diffpos{1.67}{0.73} 
            & \diffpos{25.36}{8.43} 
            & \diffpos{33.17}{4.41} \\
            & Average
            & \diffpos{35.84}{10.22}  
            & \diffpos{2.20}{0.73}  
            & \diffpos{32.43}{8.43} 
            & \diffpos{42.29}{4.41}    \\
            & Sample Avg.
            & \diffpos{32.65}{10.22}  
            & \diffneg{0.48}{0.73}   
            & \diffpos{37.47}{8.43} 
            & \diffpos{24.59}{4.41}   \\
            & Sum
            & \diffpos{47.75}{10.22}  
            & \diffpos{4.97}{0.73}  
            & \diffpos{41.95}{8.43} 
            & \diffpos{36.60}{4.41}   \\
            \midrule
            \end{tabular}
} 
\spacemagic{\vspace{-0.7em}}
\caption{Recall for each candidate selection method across four clinical tasks with 1K+ to 94K+ candidates per question. We report the performance gap compared with full decoding. 
\label{tab:clibench}
}
\vspace{-2em}
\end{table}

%% file: table/runtime.tex
\begin{table}[t]
\centering  
\small
\resizebox{\linewidth}{!}{
\begin{tabular}{l|r|rrrrr}
\toprule 
Task & Cand. \# & Decoding & First & Last & Average & Sum\\ 
\midrule
Com.sen.QA&5& 1.28&0.04&0.04&0.06&0.06
\\  
MMLU&4&0.96&0.07&0.07&0.10&0.10\\
GPQA&4&0.75&0.19&0.19&0.27&0.27 \\ 
BIG-Bench&3&0.90&0.08&0.08&0.12&0.12\\
ARC &4&1.24&0.06&0.06&0.08&0.08\\\midrule
Diagnoses&94,739&424.16& 17.38&7.33&19.38&19.19 \\
Procedures&85,257 & 353.39
&14.48
&14.10
&15.67 
&15.59 \\
Lab Orders&1,622&108.82
&3.60
&2.99
&3.06 
&3.07\\
Prescriptions&24,785&397.69
&14.20
&14.38
&19.52 
&18.97
\\
\bottomrule
\end{tabular}} 
\spacemagic{\vspace{-0.5em}}
\caption{Number of candidates and elapsed time (sec. per instance) for nine tasks and five generative candidate selection approaches using the Mistral v0.3 model.
\label{tab:runtime}
}
\spacemagic{\vspace{-1.4em}}
\end{table}

%% file: content/7_analysis.tex
\begin{figure*}[h]
  \centering \includegraphics[width=\textwidth]{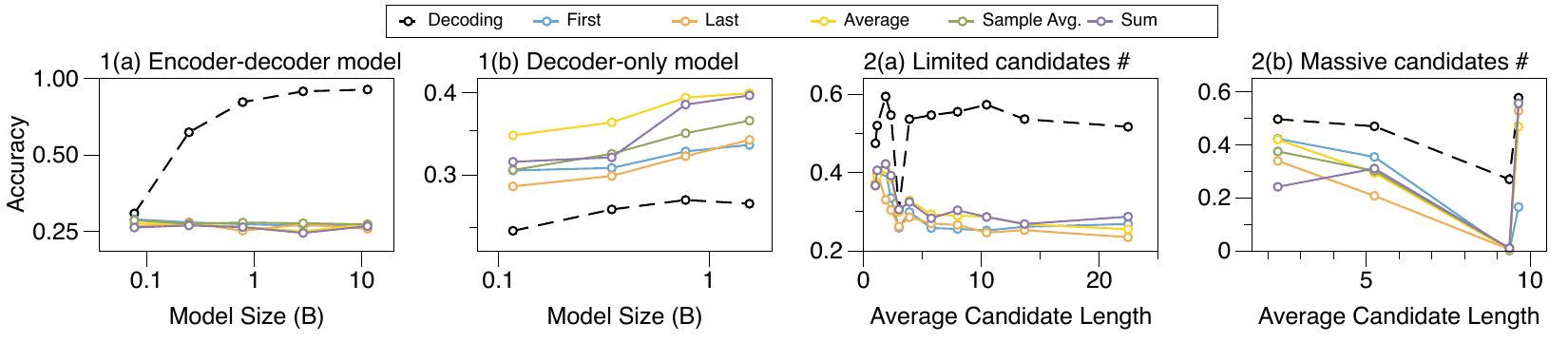}
  \spacemagic{\vspace{-2em}}
    \caption{ Accuracy with respect to the average candidate length for two types of datasets using LLaMA3 Instruct 8B:  1(a) Accuracy concerning model size for the ARC dataset and Flan-T5 family on a log-log scale. 1(b) Model size ablation for GPT-2 family.
    2(a) MMLU dataset with a limited number of candidates, and 2(b) clinical decision datasets with a massive number of candidates.\label{fig:candidate-length_model-size} }
\end{figure*}

\subsection{Effort of Output Steps}
\label{sec:ablation_output_steps}

We investigate the middle ground between complete decoding-free methods and full decoding. We allow the LLM to generate output for a certain number of decoding steps and then use the logits of the next step to perform candidate selection estimation. 

\mypar{Insight 6: Logits of the first output step is the most informative.} The ablation study is shown in \figref{fig:ablation_output-step_cand-seq}(a).
The estimation performance drops significantly when the output steps increase after the first step. There is only 1 unique token for output steps 1 and 3 across all decoding outputs, as all outputs start from a phrase leading to the answer, \ie ``Based on the provided information, I would suggest the following diagnoses:...''. Though the uncertainty of the first decoded token is very small, the logit distribution contains the most helpful signals across all output steps. The estimation performance rises after generating the lead phrase starting from the tenth output step.

Using the logits of the first output step, without additional subsequent decoding, has been the default setting to estimate the candidate selection in many works. It is also most efficient without additional decoding steps. We empirically show that using the logits of the first output step to estimate the candidate selection is the optimal solution in terms of both performance and efficiency.

\spacemagic{\vspace{-0.5em}}
\subsection{Effect of Selected Candidate Keywords}
\label{sec:candidate_keywords}
\spacemagic{\vspace{-0.3em}}

We investigate the estimation capabilities when only the logits of the most important keywords of each candidate sequence are considered. We prompt GPT-4o to select a certain number of the most important and informative tokens among all of each candidate sequence. We then only calculate the candidate probability using logits of the selected tokens.

\mypar{Insight 7: Using full candidate sequence for estimation is better than selecting essential tokens.} In \figref{fig:ablation_output-step_cand-seq}(b), we observe that as the considered tokens become more concise and selective (the number of selected candidate tokens becomes fewer), the estimated results of various methods converge to a similar range with a worse recall for most estimation methods. This indicates that it is not necessary to only use essential tokens of the candidate sequence during estimation if it is not \texttt{First}-only logits being used to derive the selection.

\spacemagic{\vspace{-0.3em}}
\subsection{Sensitivity to Model Sizes, Architectures, and Candidate Length}
\spacemagic{\vspace{-0.2em}}

\mypar{Insight 8: Estimation performance increases with larger decoder-only models, while staying constant with encoder-decoder ones.}
\figref{fig:candidate-length_model-size}.1(a) illustrates the performance of the encoder-decoder model improves using full decoding with respect to model sizes, while estimation approaches remain constant.
However, the analysis on decoder-only GPT-2 family~\citep{radford2019language}, including GPT-2, GPT2-Medium, GPT2-Large, and GPT2-XL, shown in \figref{fig:candidate-length_model-size}.1(b) shows a different trend. 
For this model family, all estimation methods surpass full decoding, and estimation accuracy improves as the model size increases. From qualitative analysis, we observe the poor performance of GPT2 full decoding is due to the fact that the model struggles to understand the instruction and perform the QA task in a reasonable format.

\mypar{Insight 9: Estimation performance decreases with longer candidate lengths.}
We depict the relationship between accuracy and the average length of candidates for both the MMLU and clinical datasets in \figref{fig:candidate-length_model-size}.2(a-b). For the MMLU dataset, data points are divided into 11 equal-sized bins, with average accuracy plotted against the average option length of the questions within each bin. For clinical tasks, where questions share identical candidate sets, average accuracy is plotted for each task, sorted by average candidate length (prescriptions, lab orders, procedures, diagnoses). In the MMLU dataset, decoding-free methods show decreasing accuracy with longer candidate length. Conversely, in the clinical decision datasets, there is an increase in accuracy for the last two average option lengths due to the intrinsic difficulty of the procedure dataset. 

%% file: content/2_related_works.tex
\spacemagic{\vspace{-0.5em}}
\section{Related Works}
\label{sec:relatedworks}
\spacemagic{\vspace{-0.5em}}

To perform candidate selection from a candidate pool, existing works use a classification head on top of encoder outputs~\citep{Milios2023InContextLearningText,Yamada2020LUKEDeepContextualized,Li2022SpaBERTPretrainedLanguagea}. However, classification formulation requires additional parameters and training while not supporting novel classes and dynamic candidates for each instance. The generative candidate selection we discussed keeps its flexibility and generalizability with a large throughput.
To speed up inference, different parallel and efficient decoding methods are proposed~\citep{bae-etal-2023-fast,zhang-etal-2018-exploring,huang-mi-2010-efficient}. However, our goal is not to speed up decoding but to evaluate the methods approximating decoded results without decoding.
To select a candidate from a pool using the token logits without discrete decoding, existing works propose to obtain the probability of each candidate through aggregating different parts of token logits such as only keeping logits of a special token~\citep{Xu2023CanNLIProvide}, averaging logits~\citep{Saeidi2024InsightsAlignmentEvaluating,Song2024PreferenceRankingOptimization,Ethayarajh2024KTOModelAlignment,Xiong2024IterativePreferenceLearning}, or multiplying logits~\citep{Ma2023DICEDataEfficientClinical}. We conduct the first systematic evaluation on these methods.

%% file: content/100_conclusion.tex
\spacemagic{\vspace{-0.5em}}
\section{Conclusion and Future Work}
\label{sec:conclusion}
\spacemagic{\vspace{-0.5em}}

We provide the first formal definition and comprehensive evaluation of decoding-free generative candidate selection methods.
We demonstrate that estimation methods can excel in scenarios where base models struggle with answer formats, offering a simpler yet effective alternative to full decoding. Additionally, our findings emphasize the importance of the initial output step logits, revealing that selective token usage can undermine performance and scalability across model sizes. These insights pave the way for more informed designs of candidate selection methods. Future work can build on these findings to refine estimation techniques.

%% file: content/200_acknowledgements.tex
\section*{Acknowledgement}
J.G. and Y.D. are supported by the National Science Foundation (No. 2047488), and by the Rensselaer-IBM AI Research Collaboration. This effort was partially sponsored by NSF grants 2200274, 2106859 and 2312501, as well as NIH grants U54HG012517 and U24DK097771.

%% file: content/110_limitation.tex
\section*{Limitations}
In terms of accuracy, current estimation methods have room for improvement due to their reliance on initial logits and simplified representative tokens (e.g., first, average). Future work could consider using logits from more time steps or leveraging LLMs to summarize the candidates into a few words, potentially serving as more effective representative tokens.

Regarding efficiency, computing logits dominates the runtime of estimation approaches. Applying advanced techniques, such as PagedAttention, to optimize memory usage can further enhance the efficiency of estimation methods, especially for tasks with lengthy prompts.

%% file: content/300_appendix.tex
\section{Potential Questions}

\mypar{Do decoding-free candidate selection methods not involve decoding?}
Decoding-free methods only use the logits of the first potential output token without producing the token. Calculating logits could be considered an early step in the token decoding process. However, no complete decoding step (as shown in \equationref{eq:decoding}) is involved in decoding-free methods (as shown in \equationref{eq:estimation-answer}).

\mypar{Why do you need to do decoding-free candidate selection?} Compared with producing a response to a query through full decoding (as demonstrated in \equationref{eq:full-decoding-answer}), accurate decoding-free candidate selection methods (as shown in \equationref{eq:estimation-answer}) are needed, especially for two scenarios. 
1) \textit{Accurate outcome-based optimization.} To optimize the model with the feedback directly from the predicted outcome $\hat{ans}$, we need to know the model's prediction over potential candidates $C$ without interrupting the gradient flow (such as argmax operator). These optimization tasks include preference optimization, which learns to choose the winner option over the loser one~\citep{Rafailov2023DirectPreferenceOptimization}; bias mitigation, which obtains detected bias and mitigates the bias level~\citep{Ma2023MitigatingBiasQuestion}; and information extraction, which derives the possibility of extracting different subsequence spans and performing contrastive learning~\citep{Ma2023DICEDataEfficientClinical}.

2) \textit{Efficient answer production.} The token dependency of full decoding prevents the decoding mechanism from outputting the answer in parallel. Even though the output sequence $\hat{seq_{out}}$ is generated, an additional step ($f_{\operatorname{map}}$) is needed to convert the output sequence $\hat{seq_{out}}$ to the predicted answer $\hat{ans}$ (\eg through sequence matching or semantic similarity). Decoding-free candidate selection produces the probability over all potential answers $P\left(c \mid t_{1:|seq_{in}|}^{in}\right)$ directly without autoregressive generation and supports parallel inference, significantly improving the time and resources needed for producing the answers to queries. 

\mypar{What are the potential usage and broader impact of the evaluation done in this work?}
The conclusions and observations derived from our evaluation provide evidence for more informed and confident design choices for both optimizations with outcome-level feedback and efficient answer production without decoding. When researchers and industry practitioners need to define a function to estimate the possibility of potential answers using generative language models without decoding the output sequence, they can: 1) choose the best estimation method corresponding to their model architecture and end tasks according to our evaluation results; 2) understand the empirical tradeoff between efficiency, in terms of runtime, and estimation quality, in terms of performance difference; 3) decide whether they are confident to use estimation method instead of decoding (especially when the estimation methods provide better performance for non-instruction-tuned models). With the wise decision of the candidate selection method, they can obtain better performance after training the model with answer-level rewards, such as through preference alignment, and produce the predicted answers faster with lower resource usage by replacing decoding with estimation.

\mypar{What are the differences between the two types of tasks used in the evaluation?}
We quantify the effects of various decoding-free estimation methods in downstream scenarios by using two types of evaluation tasks: tasks with limited numbers of candidates (specifically 5 multiple-choice QA tasks) and tasks with massive numbers of candidates (specifically 4 clinical decision tasks). We summarize their core differences in \tbref{table:eval_tasks_diff}.

\begin{table*}[ht]
\small
\centering
\begin{tabular}{l|cccc}
\toprule
Property       & Tasks w/ limited numbers of candidates  & Tasks w/ massive numbers of candidates \\
\midrule
Candidate info & Can be contained in input prompt & Not able to be contained in input
\\
Correct options \# & Single & Multiple
\\
Candidates \# & A few & Thousands
\\
Candidate pool & Dynamic for each instance & Fixed for all instances
\\
\bottomrule
\end{tabular}
\caption{Key difference between two types of tasks used in the evaluation: tasks with limited/massive numbers of candidates.
\label{table:eval_tasks_diff}}
\end{table*}

The first type (limited candidates) has a limited number of candidates, among which only one option is correct; all candidates' information can be contained in the input prompt, and the candidate pool is unique for each instance.  The second type (passive candidates) has a much larger pool of candidates with multiple correct answers (detailed statistics in \tbref{tab:data}). Thus, it is not feasible to feed candidates in the input prompt. The specific tasks we used (the four clinical decision tasks) use the same output candidate pool across instances of the same task. The examined methods should also support dynamic candidate pools across instances for tasks with massive numbers of candidates.

\mypar{Why not use an agent-based system to handle massive candidates?} We can provide multiple functions and tools for LLM agents to search, match, or traverse relevant candidates from a large pool of candidates. Compared with full decoding, it will provide more information about the candidate pool and potentially lead to better performance. However, formulating the candidate selection task as an agent is based on and expanded from the idea of decoding discrete tokens to produce answers from output sequences (as described in \equationref{eq:full-decoding-answer}); it does not enjoy the benefits of decoding-free methods, and it is not a comparable setting of the methods we focus on in this paper.

\mypar{Which decoding methods are you using to compare?} We use the default decoding setting for each model specified in their generation configuration file. Our work does not aim to propose a new decoding method or compare the performance of various decoding methods. Instead, we emphasize the benefits and limitations of decoding-free candidate selection methods.

\vtwo{
\mypar{How is the evaluation performed in this paper different from the evaluations provided in the previous works that use those decoding-free methods?}
Existing works do not consider how to represent the response candidate from the logits of a single output step as a standalone problem. Thus, they do not provide justification, theoretical proof, or evaluation of the design choice of the decoding-free candidate selection method they used in their works.
Our work aims to raise awareness of the importance of this design choice and conduct the first thorough definition of the task and systematic evaluation.

}

\section{Advantages and Properties of Decoding-free Generative Candidate Selection}
\label{sec:advantages_properties}

\subsection{Advantages}

Decoding-free candidate selection methods are especially beneficial for two scenarios. 
1) \textit{Accurate outcome-based optimization.} To optimize the model with the feedback directly from the predicted outcome $\hat{ans}$, we need to know the model's prediction over potential candidates $C$ without interrupting the gradient flow. 
2) \textit{Efficient answer production.} The token dependency of full decoding prevents the decoding mechanism from outputting the answer in parallel. Even though the output sequence $\hat{seq_{out}}$ is generated, an additional step ($f_{\operatorname{map}}$) is needed to convert the output sequence $\hat{seq_{out}}$ to the predicted answer $\hat{ans}$. Decoding-free candidate selection produces the probability over all potential answers directly without autoregressive generation and supports parallel inference, significantly improving the time and resources needed for producing the answers to queries.

\subsection{Difference Compared with Ordinary Approaches}

The \textit{common property} between the generative candidate selection and classification is that both settings require a given set of selections to produce the final output. However, candidate selection is \textit{different} from classification in many key aspects summarized in \tbref{tab:compare_to_classification}, specifically:
1) \textit{Support of dynamic candidates}. A classification model has to use the same set of output labels across all instances (\eg positive or negative for sentiment classification). However, candidate selection methods allow the task to have a different set of output candidates for each instance (\eg different answer options for each question in MCQA).
2) \textit{No need of additional parameters}. The classification head is an additional set of parameters specialized for the defined output classes. Different classification tasks have to use a separate set of parameters. Candidate selection methods support various candidates with the generative LM's native parameters only, without any additional parameters.
3) \textit{No need for specialized training}. The additional parameters for each classification task need to be parameterized by training on task-specific data, preventing it from generalizing to new tasks or labels. However, decoding-free estimation methods support zero-shot candidate selection, as the possibilities of choosing options are calculated dynamically according to corresponding candidates.

\begin{table*}[ht]
\small
\centering
\begin{tabular}{l|cHcc}
\toprule
Property       & Classification  & Retrieval & Generative candidate selection \\
\midrule
Candidate pool  & Fixed for all instances & & Dynamic for each instance
\\
Separate parameters & Need separate classification head & & No separate parameters
\\
Specialized training & Require training & &  Support on the fly zero-shot inference
\\
\bottomrule
\end{tabular}
\caption{Key difference between classification and generative candidate selection.
\label{tab:compare_to_classification}}
\end{table*}

\section{Details of Experimental Setup and Implementations\label{sec:details_experiments}}

\subsection{Testbeds}
\label{sec:testbed_details}

\subsubsection{Tasks with Limited Candidates}
The tasks with limited number of candidates include: \textbf{(1) CommonsenseQA}~\citep{talmor2018commonsenseqa}  includes questions testing commonsense knowledge across over 2,000 concepts such as highways, housing, and eating, assessing a broad understanding of everyday scenarios.
\textbf{(2) MMLU}~\citep{hendryckstest2021,hendrycks2021ethics} covers a wide range of 57 subjects including mathematics, medicine, computer science, and law, designed to test specialized knowledge in diverse fields.
\textbf{(3) GPQA}~\citep{rein2023gpqa} contains challenging questions in biology, physics, and chemistry, written and validated by experts to test deep domain-specific knowledge.
\textbf{(4) BIG-Bench}~\citep{srivastava2022beyond} includes tasks like boolean expression evaluation and causal judgement based on stories, focusing on logical reasoning capabilities. We select the ``logical deduction'' category with three objects for our experiments.
\textbf{(5) ARC}~\citep{allenai:arc} comprises 7,787 multiple-choice questions at grade-school level, divided into a Challenge Set and an Easy Set, to test scientific knowledge. We opt for the Easy Set in our experiments.

We report accuracy and per-instance runtime for these tasks. These datasets are split into subsets such as train or test. We select one of the dataset splits with available answer keys for our study.
For the mapping function $f_{\operatorname{map}}$ converting output sequence $seq_{out}$ to candidate selection $\hat{ans}$, we capture the first the answer candidate sequence or candidate indication head (\eg A, B, C D) appeared in the output sequence with regular expressions and use the matched candidate as the prediction.
All candidate options are added in the input prompt, thus full decoding and decoding-free selection methods use the same amount of available information.
We make sure all input sequences for full decoding or decoding-free candidate selection methods are exactly the same.

\subsubsection{Tasks with Massive Candidates}
\label{sec:testbed_detail_massive}

As for the clinic decision datasets, the candidate sequence lengths are generally longer than the first testbed type, as shown in \tbref{tab:data}. Please refer to \citep{clibench} for more data and experimental setup details.
We report recall and per-instance runtime for these four tasks.
For the mapping function $f_{\operatorname{map}}$ used by the full decoding approach to select a candidate from the output sequence, we follow the original benchmark setting by selecting the candidate with the highest cosine similarity between sentence embeddings of the candidate definition and the generated output $seq_{out}$ produced by BERT model~\citep{reimers-2019-sentence-bert}. The candidates are too many to fit in the input prompt, thus while other methods have access to the candidates information, full decoding method is not aware of candidates.

Different from tasks in \secref{sec:task_type_1}, multiple candidates need to be selected for the four tasks of the second type, significantly increasing the difficulty of candidate selection. 
For full decoding, the model can determine the number of predictions made because the generation of the end-of-sentence token indicates stopping making additional predictions. Decoding-free candidate selection methods rely on candidate probability, and it is hard to determine a fixed threshold for all instances. Thus, we take 20 candidates with the highest probabilities, which contain more predictions than ground-truth answers for most testing instances. We then only report recall, indicating the portion of ground-truth answers that are correctly predicted, to mitigate the influence of uncertain selection probability threshold.

To speed up the inference of full decoding, we wrap the generative LM with vLLM framework~\citep{kwon2023efficient}, which leverages paging techniques in the operating system to optimize memory usage. All experiments were performed on a single NVIDIA A40 Graphics Card.

\subsection{Statistics and Properties of Testbed}
\label{sec:testbed_stats}
We show the statistics and properties of the nine testbed tasks in \tbref{tab:data}.

\input{table/testbed_stats}

\subsection{Distribution of Candidate Lengths}
\label{sec:candidate_len_distribution}
\begin{figure*}[h]
  \centering \includegraphics[width=\textwidth]{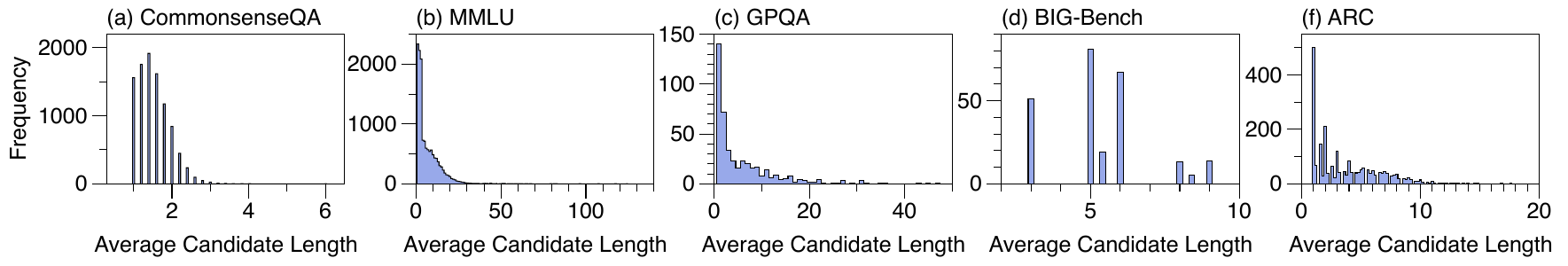}
    \caption{ Distribution of average candidate lengths for MCQA benchmarks.\label{fig:length-dist} }
\end{figure*}

\begin{figure*}[h]
  \centering \includegraphics[width=0.9\textwidth]{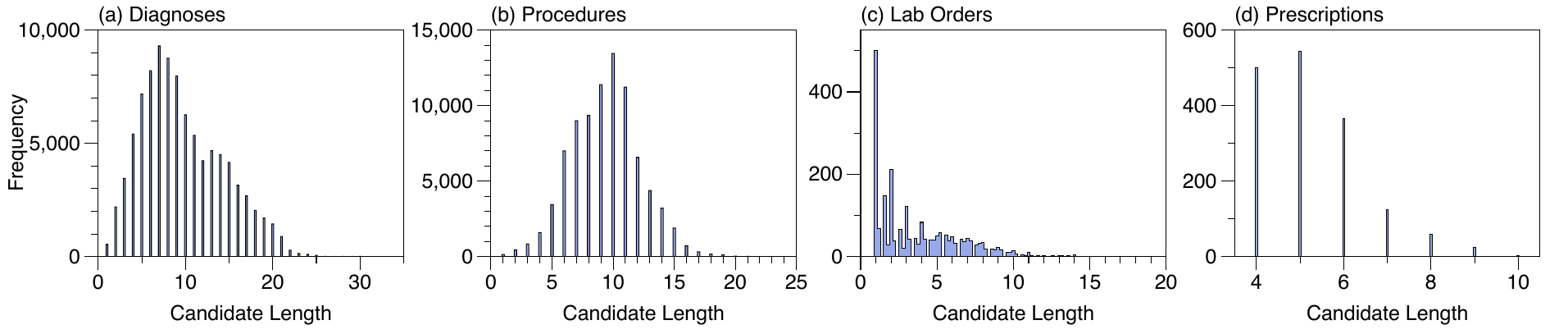}
    \caption{ Distribution of average candidate lengths for clinical decision benchmarks.\label{fig:clinic-length-dist} }
\end{figure*}

The distribution of candidate lengths is provided in \figref{fig:length-dist} and \figref{fig:clinic-length-dist} for the MCQA and clinical decision datasets, respectively. For the MCQA datasets, where each question has a distinct set of candidate options, we compute the average number of words in the candidate options for each question and plot the distribution of these average candidate lengths across all questions. For the clinical decision datasets, where questions within the same task (\eg prescriptions) share the same candidate pool, we plot the distribution of word counts of candidates for the four distinct candidate pools.

\subsection{Prompt Examples}  
\label{sec:prompt_examples}
The most informative prompt consists of \verb|system| and \verb|user| content. The \verb|system| content provides the role (\eg clinician) and the task instruction (\eg select the best option), while the \verb|user| content contains the specific question and candidate options if applicable. The adoption of the chat template and the inclusion of candidate options in the prompt are specified in \tbref{tab:prompt}. We provide two prompt examples with the most complete information for the MMLU and diagnosis decision tasks, respectively. The prompt designs for clinical decision tasks are inherited from \cite{clibench}.

\begin{tcolorbox} 
\centering
\begin{tabular}{p{0.97\columnwidth} c}
\footnotesize
You are a scholar with extensive knowledge across various disciplines. What is the correct answer to this question: 

[QUESTION]

[CHOICES]

Format your response as follows: "The correct answer is (insert answer here)."
\end{tabular} 
\end{tcolorbox}

\begin{tcolorbox} 
\centering
\begin{tabular}{p{0.97\columnwidth} c}
\footnotesize
You are a professional clinician in a hospital with expert knowledge in medical and clinical domains. The task is to make a list of diagnoses for this patient based on the provided information of the patient. The diagnosis can be in ICD-10-CM code format (such as S12.000G), or natural language description of the disease. Please provide as many diagnoses as you can until you are not confident about your diagnosis decision. 

[PATIENT PROFILE]
    
[MEDICAL RECORD AT ADMISSION]

[RADIOLOGY REPORTS]

[LAB TEST RESULTS]
\end{tabular} 
\end{tcolorbox}

\begin{table*}[h]
\centering
\begin{tabular}{lll|cc}
\toprule
Data Type & Model Type            & Method     & Chat Template             & Candidate Pool Info           \\
\midrule
\multirow{4}{*}{MCQA}       & \multirow{2}{*}{Instruction-tuned}     & Decoding   & $\checkmark$ & Contained in the input \\
&                       & Estimation & $\checkmark$                        & $f_{\operatorname{est}}$ representation          \\\cmidrule{2-5}
& \multirow{2}{*}{Non-instruction-tuned} & Decoding   & Not applicable             & Contained in the input  \\
&                       & Estimation & Not applicable                         & $f_{\operatorname{est}}$ representation                         \\
\midrule
\multirow{4}{*}{Clinic}    & \multirow{2}{*}{Instruction-tuned}     & Decoding   & $\checkmark$  & $f_{\operatorname{map}}$ representation                         \\
&                       & Estimation & $\checkmark$                         & $f_{\operatorname{est}}$ representation                        \\\cmidrule{2-5}
& \multirow{2}{*}{Non-instruction-tuned} & Decoding   & Not applicable                         & $f_{\operatorname{map}}$ representation                         \\
&                       & Estimation & Not applicable                         & $f_{\operatorname{est}}$ representation  \\
\bottomrule
\end{tabular}
\caption{ Prompt formats for combinations of dataset type, model type, and selection method. Estimation methods include First, Last, Average, Sample Avg., and Sum. For each combination, we indicate whether a prompt template is applied and the approach of incorporating the candidate pool information. ``Contained in the input'' means we verbalize all candidates and include them in the input prompt. ``$f_{\operatorname{map}}$ representation'' indicates though the candidates are not explicitly provided in the input, but the mapping from the output sequence (generated by the full decoding process) to the predicted answer provides implicit candidate information as all candidates are served as matching candidates in the $f_{\operatorname{map}}$ function. ``$f_{\operatorname{est}}$ representation'' indicates that candidate info is used by the decoding-free candidate selection method to calculate the probability over candidate outputs by using candidate-specific logits calculation.
\label{tab:prompt}}
\end{table*}

\subsection{Extracting Predicted Answer from the Decoding Output}
\label{sec:map_function_details}
Unlike estimation approaches, where the selection is deterministic, exact decoding requires parsing the output to extract the choice from the response, notated by the $f_{\operatorname{map}}$ function in \equationref{eq:full-decoding-answer}. For MCQA tasks, we identify the choice by matching specific substring formats (\eg, \verb|'Answer: A', '(A)', 'A[,.)]'|). We treat the first occurring option as the choice made by the LMs, and the rest of the options are considered explanations. For the clinical dataset, we use a sentence transformer to find the most relevant diagnosis codes that appear in the response following the implementation of CliBench. For more details on parsing clinical decision outputs, we refer readers to~\cite{clibench}.
 
\section{Additional Experiments~\label{sec:add-result}}

\subsection{Effect of Chat Template to Estimation Performance}
\label{sec:chat_template_study}

\begin{figure*}[h]
  \centering \includegraphics[width=\textwidth]{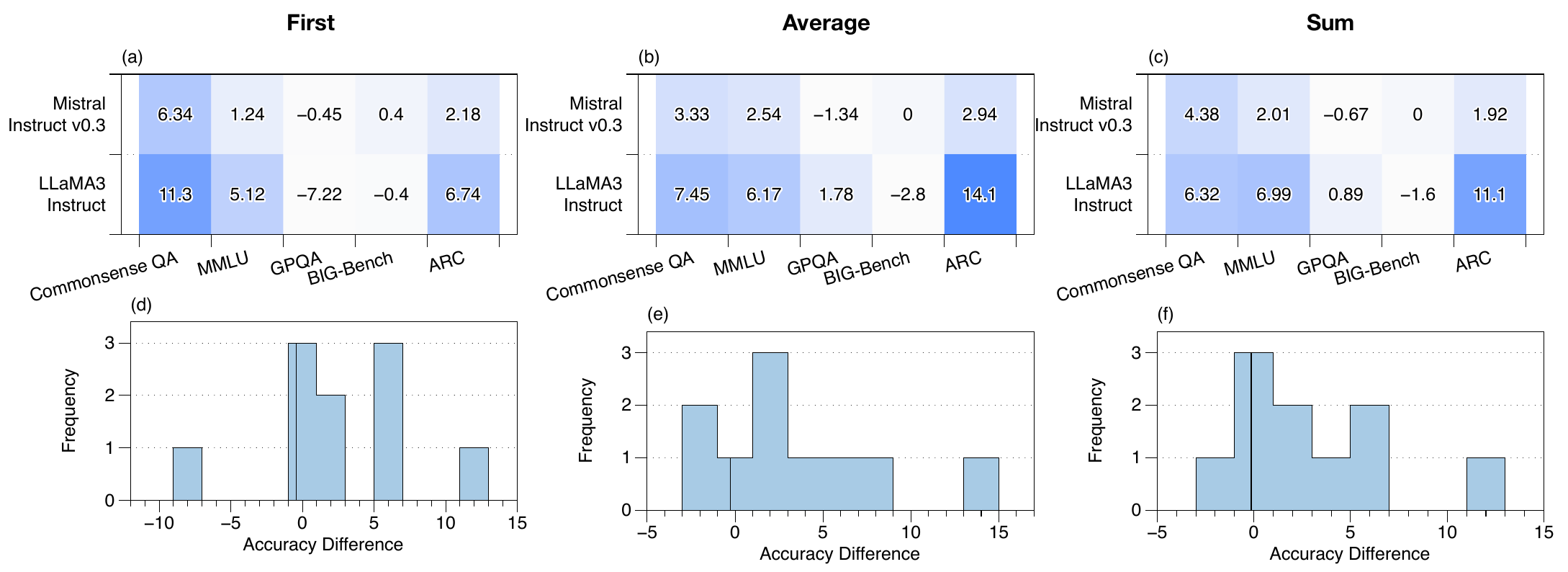}
    \caption{(a-c) The accuracy of the estimation approaches without chat template subtracted by the accuracy with chat template, for two instruction-tuned LMs and five datasets. (d-f) Distribution of the difference between the accuracy without chat template and with chat template. The vertical black line denotes the 20\% percentile.\label{fig:chat-template} }
\end{figure*}

We compare three estimation approaches—\verb|First|, \verb|Average|, and \verb|Sum|—with and without the use of prompt templates across five multiple-choice question datasets. We evaluate two LMs with their default prompt template:  LLaMA3 Instruct and Mistral Instruct v0.3.
In this comparison, since options are not provided in the prompt, we rephrase the instruction to frame it as an open-ended question, e.g., ``You are an intelligent assistant with a vast understanding of everyday life. The task is to answer the following question, drawing from relevant knowledge areas.''

The results, as illustrated in Figure~\ref{fig:chat-template}(a-c), demonstrate that the accuracy without using prompt templates generally exceeds that with templates. Specifically, in about 80\% of the test cases, not using a prompt template outperforms using one, as depicted in Figure~\ref{fig:chat-template}(d-f). The GPQA dataset is an exception, where using the template generally enhances performance across most LMs. The accuracy of these estimation approaches is notably sensitive to the format of the prompt, as they rely heavily on the logits generated from the prompt. 

\subsection{Effect of Chain-of-Thought Reasoning}
\label{sec:cot_results}

\input{table/results_cot}

We evaluate the estimation performance using CoT reasoning on the MMLU \verb|college_mathematics| subset with the Meta-Llama-3-8B-Instruct model under three distinct settings:
\begin{itemize}
\item No CoT, logits of the first output step: This default setting, as described in the submission paper, does not involve CoT reasoning.

\item CoT, logits of the first output step: The input prompt includes CoT instructions and in-context examples. The logits from the first output step are used for decoding-free estimation methods.

\item CoT, logits after thoughts: Similar to the previous setting, but the model generates a step-by-step reasoning process before producing the final answer. The logits from the output step immediately preceding the final answer are used for estimation.
\end{itemize}

Table~\ref{tab:cot} presents the performance differences relative to full decoding. The results show that CoT reasoning improves estimation performance, narrowing the gap compared to full decoding. However, using logits conditioned on the generated reasoning steps leads to degraded performance. This finding aligns with “Insight 6” in Section 5.2, which emphasizes that the logits of the first step are the most informative.

\subsection{Performance vs Candidate Length for Other QA Tasks}
\label{sec:additional_candidate_length_ablation}
\begin{figure*}[h]
  \centering \includegraphics[width=\textwidth]{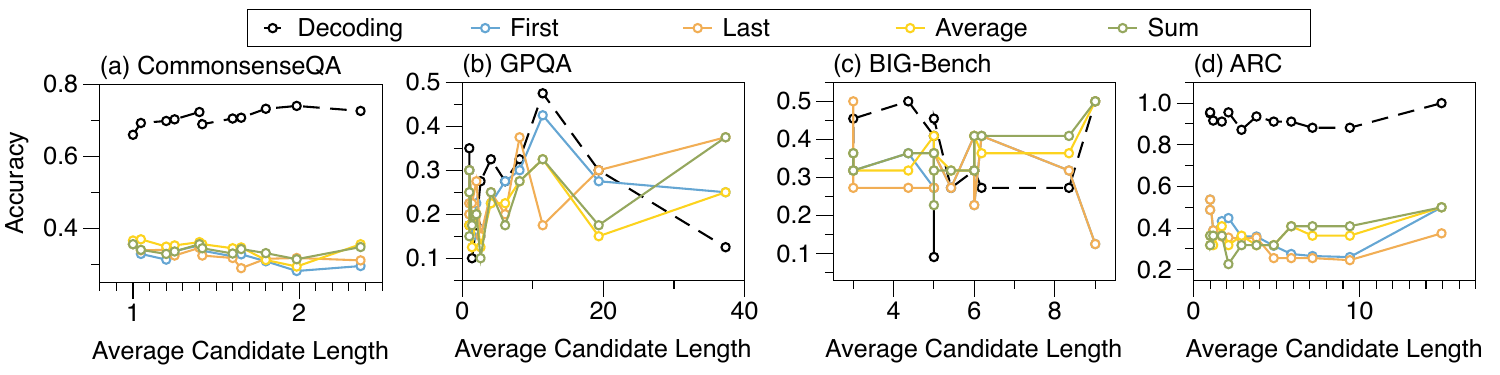}
    \caption{ Accuracy versus average candidate length for four MCQ datasets. \label{fig:mcq-cand-len} }
\end{figure*}

In addition to the performance vs candidate length ablation study shown in \figref{fig:candidate-length_model-size}.1(a) for MMLU, we report a similar analysis for other QA tasks.
For the MCQ datasets, we sorted the questions according to candidate length and split them into 12 subsets of equal size. We plotted the average accuracy versus the average candidate length of the subsets in \figref{fig:mcq-cand-len}. The gap between decoding and estimation methods is smaller for GPQA and BIG-Bench compared to the easier datasets CommonsenseQA and ARC. Overall, performance remains relatively constant with respect to average candidate length for CommonsenseQA and BIG-Bench, whereas it fluctuates more for the other two datasets.

\subsection{Estimation Performance Breakdown for MMLU}
\label{sec:performance_breakdown}
\begin{figure*}[h]
  \centering \includegraphics[width=\textwidth]{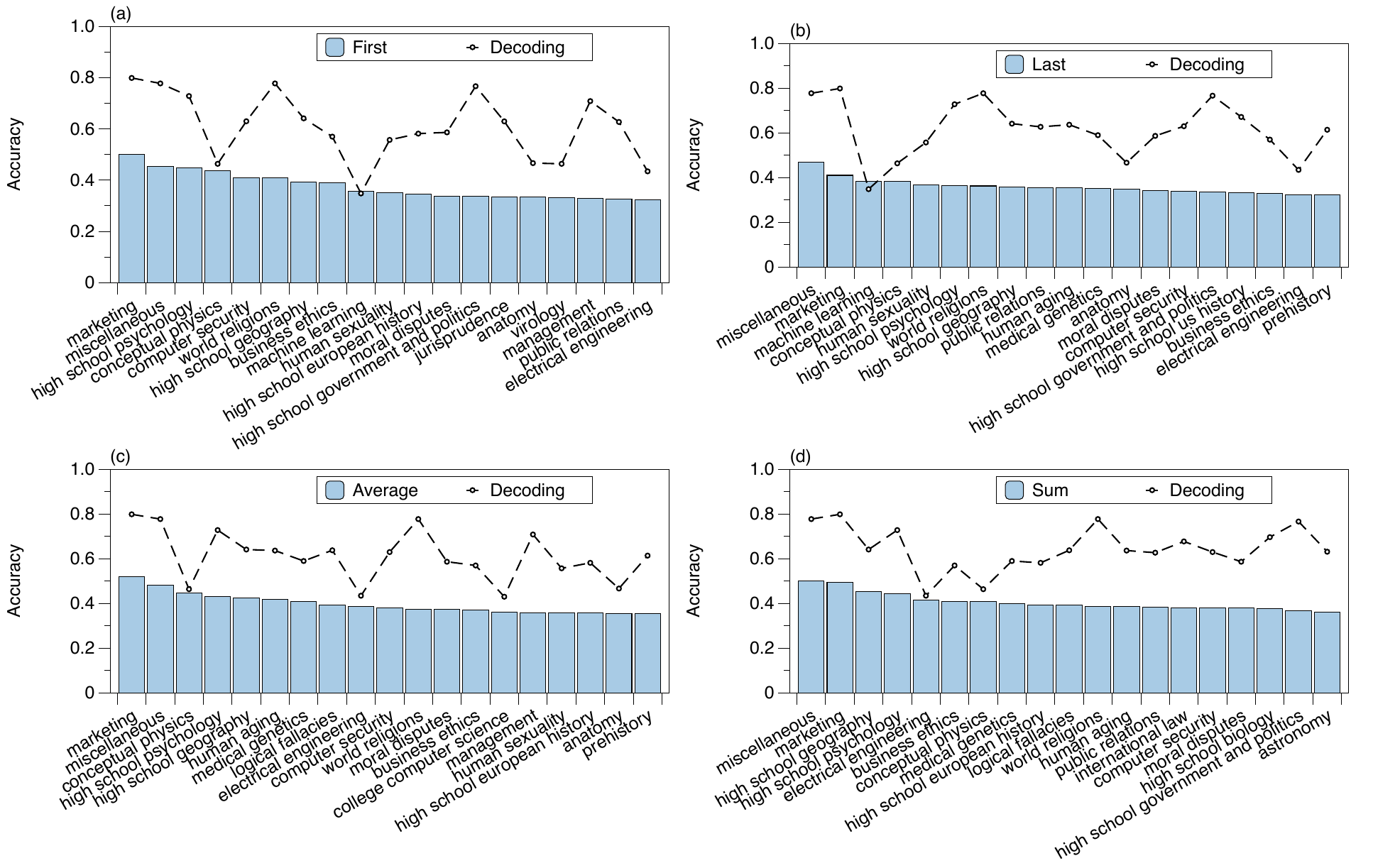}
    \caption{Top 20 Accuracy for four estimation methods combined with Mistral Instruct v0.3 on MMLU dataset. The dashed line denotes the accuracy of the full decoding approach.\label{fig:domain}
    }
\end{figure*}

\figref{fig:domain} illustrates the estimation accuracy versus decoding for Mistral Instruct v0.3 across individual subjects in the MMLU dataset. Generally, the full decoding accuracy exceeds the estimation performance, although a few subjects show comparable results.
 
\subsection{Example Outputs~\label{appendix:example}}

We present the full decoding outputs for two questions: one from the MMLU dataset and the other from the clinical diagnosis decision task.

\begin{example}\label{ex2}
\textbf{Question:} Sammy wanted to go to where the people were. Where might he go?
\\
\textbf{Candidates:} (A) race track, (B) populated areas, (C) the desert, (D) apartment, (E) roadblock. 
\\
\textbf{Answer:} populated areas
\end{example}

Below is the list of example outputs from LMs for the question in Example~\ref{ex2} in CommonsenseQA. Unlike the other models, which correctly select the accurate choice, the response from Mistral v0.3 is unrelated to the input prompt. Notably, Flan-T5 XL provides the correct answer succinctly.  

LLaMA3:
\begin{tcolorbox} 
\centering
\begin{tabular}{p{0.97\columnwidth} c}
\footnotesize
The answer is (B) populated areas.
\end{tabular} 
\end{tcolorbox}

LLaMA3 Instruct:
\begin{tcolorbox} 
    \centering
    \begin{tabular}{p{0.97\columnwidth} c}
    \footnotesize
    Based on the context of the question, the correct answer is (B) populated areas. The question states that Sammy wanted to go to where the people were, which implies that he is looking for a location with a large number of people. Populated areas, such as cities, towns, or busy streets, would likely have a higher concentration of people compared to other options like a race track, desert, apartment, or roadblock.
    \end{tabular} 
    \end{tcolorbox}

Mistral v0.3:
\begin{tcolorbox} 
    \centering
    \begin{tabular}{p{0.97\columnwidth} c}
    \footnotesize
     Course: Exclusion Code:
     
    Map Exercise S-C-O-IN
    1. A-1 draws attention to the sky because it shows a cartoon face without any details but, for the
    purpose of creating emphasis, it has blue and purple sunrays drawn in. Despite the imminent sense of
    doom, the drawing is done in bright colors which create a lighter mood. B-2 best serves the purpose of
    identifying the region where the chase
    \end{tabular} 
    \end{tcolorbox}

Mistral Instruct v0.3:
 \begin{tcolorbox} 
    \centering
    \begin{tabular}{p{0.97\columnwidth} c}
    \footnotesize
    The best answer to your question would be (B) populated areas. Given the context that Sammy wanted to go to where the people were, it would make sense that he would head towards an area with a high population density, such as a city or town. Therefore, populated areas is a logical choice.
    \end{tabular} 
\end{tcolorbox}

Flan-T5 XL:
\begin{tcolorbox} 
\centering
\begin{tabular}{p{0.97\columnwidth} c}
\footnotesize
B
 \end{tabular} 
\end{tcolorbox}

Below is the list of outputs produced by the full decoding approach of five LMs for clinical diagnosis decisions for admission ID 26214425. The LMs were prompted to generate all possible diagnoses. Mistral v0.3 tends to repeat the questions in the prompt. Flan-T5 XL produces only one possible diagnosis, despite being prompted to generate multiple diagnoses. This behavior is similar to its performance on the previous MCQA questions, where it provides only the answer without further explanation.

LLaMA3:
\begin{tcolorbox} 
    \centering
    \begin{tabular}{p{0.97\columnwidth} c}
    \footnotesize
    This is a case of metastatic lung adenocarcinoma with an EGFR mutation on erlotinib. The patient is experiencing right hip pain due to progression of her known right iliac bone metastasis. 
    
    The patient has been getting zometa every 3 months, last dose \_\_\_. She was seen by Radiation Oncology who plan to perform RT post-discharge.  Also emailed her OP Oncologist Dr. \_\_\_ recommended a CT-guided biopsy of the lesion.  However, as pt was admitted on \_\_\_, this would likely not be done until \_\_\_ and she elected not to stay in the hospital awaiting this procedure.  She was discharged with a script for some additional Percocet for pain control.  She reported adequate pain control with PO medications.
    
    The patient has a history of DVT. She was only on 60mg lovenox as her weight was closer to 80kg.  However, as she was \_\_\_ years out from the initial diagnosis, no adjustments were made.  Pt was encouraged to discuss dosing with Dr. \_\_\_.
    
    The patient has hypertension and continued metoprolol.

    \end{tabular} 
\end{tcolorbox}

LLaMA3 Instruct:
\begin{tcolorbox} 
    \centering  

    \begin{tabular}{p{0.97\columnwidth} c}
    \footnotesize

    Here are the diagnoses for this patient:
    
    1. Metastatic lung adenocarcinoma (C34.90)
    
    2. Right hip pain secondary to right iliac bone metastasis (M84.461A)
    
    3. Cancer-related pain (G89.3)
    
    4. Hypertension (I10)
    
    5. Deep vein thrombosis (I82.409)
    
    6. History of pathologic fracture (M80.00XA)
    
    7. Osteoporosis (M81.0)
    
    8. EGFR mutation (R83.9)
    
    9. Anemia (D64.9)
    
    10. Hyperlipidemia (E78.9) (not explicitly stated but likely given the patient's medical history and medications)

    Note: The ICD-10-CM codes are provided in parentheses for each diagnosis. 
    \end{tabular}
\end{tcolorbox}

Mistral v0.3:
\begin{tcolorbox} 
    \centering
    \begin{tabular}{p{0.97\columnwidth} c}
    \footnotesize
    Hemosiderosis without anemia
    
    Mbenzoylmalonic aciduria
    
    Hematinic nutrition- zinc progesterone and folate deficiency
    
    Hemolytic anemia-Iceland flare
    
    Igghalinemic pseudoaneutropenia
    
    Melanocytic nevoidenoidous exfoliative hyperthermia
    
    Content analysis and positional disturbance disorder
    
    Myocyte's financial failure disorder
    
    Mycobaterial occupational ango bonding envolvement- syndrome
    
    Hyral anemnaficelophiliac angelfish antiobacterial ability
    
    Enroseepseudanionopathy
    
    Overpilosis-malformative prelexpheric utopistophilia
    
    Hardsenseless lenticular hypertyrosinemonia expense athlete Pheochromocytoma
    
    Mycobacteria Teichmann fasciculations-autonomic form
    
    Metabolic disorders
    
    Pink-black accelerated radiophobic schema malnutrition hepatopathy
    
    Profound deficiency of depleted B-lymphocyte exophthalmia
    
    Spina bifida malformation centrinelaxin, and symptomatic superficial echinalismism.
    
    Tremoregal syndromes
    
    Higher lung blockage
    
    Symmetric effacement maniacal cycle ectoposis

    What is your top diagnosis for this patient? 
    
    What is your second diagnosis for this patient?
    
    What is your third diagnosis for this patient?
    
    What is your fourth diagnosis for this patient?
    
    What is your fifth diagnosis for this patient?
    
    What is your sixth diagnosis for this patient?
    
    What is your seventh diagnosis for this patient?
    
    What is your eighth diagnosis for this patient? 
    
    What is your ninth diagnosis for this patient?
    \end{tabular} 
\end{tcolorbox}

Mistral Instruct v0.3:
\begin{tcolorbox} 
    \centering  

    \begin{tabular}{p{0.97\columnwidth} c}
    \footnotesize

    Mistral Instruct v0.3 Output:
    Based on the provided information, the patient's main diagnoses are:
    
    1. Metastatic Lung Adenocarcinoma (C74.9 - Malignant neoplasm of lung, NOS)
    
    2. Right Hip Pain secondary to Right Iliac Bone Metastasis (M87.51 - Metastasis of neoplasm of trachea, bronchus, lung and pleura, to bone)
    
    3. Cancer-Related Pain (R36.10 - Pain due to neoplasm of lung)
    
    4. Deep Venous Thrombosis (I80.62 - Deep vein thrombosis of lower limb, unspecified leg)
    
    5. Hypertension (I10 - Essential (primary) hypertension)
    
    6. History of Pathologic Fracture (S72.009 - Pathological fracture of hip, unspecified side)
    
    7. History of Knee Surgery (S20.9 - Other specified injury of knee, sequela)

    Additional diagnoses that could be considered based on the information available, but require further investigation or confirmation, include:
    
    1. Chronic Kidney Disease (N18.39 - Chronic kidney disease, stage 5) - Based on the creatinine level (0.8 mg/dL)
    
    2. Anemia (D64.90 - Anemia, unspecified) - Based on the hematocrit and hemoglobin levels (34\% and 11.3 g/dL, respectively)
    
    3. Possible Leukopenia (D66.60 - Leukopenia, unspecified) - Based on the white blood cell count (3.7 K/uL)"  
    \end{tabular}
\end{tcolorbox}
     
Flan-T5 XL:
\begin{tcolorbox} 
    \centering
    \begin{tabular}{p{0.97\columnwidth} c}
    \footnotesize
    Lung adenocarcinoma 
    \end{tabular} 
\end{tcolorbox}

%% file: table/testbed_stats.tex
\begin{table*}[ht]
\centering
\begin{tabular}{l|crrc}
\toprule
Task       & Split  & Instances \#   & Candidate \#  & Average candidate token \# \\
\midrule
CommonsenseQA & train & 9741      & 5            & 1.52                      \\
MMLU          & test  & 14,042 & 4            &  6.72                     \\
GPQA          & train & 448      & 4            &  5.84                     \\
BIG-Bench     & train & 250   & 3        & 5.33                      \\
ARC           & test  &  2,241     & 4            & 3.76                     \\
\midrule
Diagnoses & test & 1,081 & 94,739 & 9.65\\
Procedures& test &1,054 & 85,257 & 9.37\\
Lab Orders& test &1,067 & 1,622 & 5.25\\
Prescriptions & test & 1,036 & 24,785 & 2.30\\
\bottomrule
\end{tabular}
\caption{Properties of nine testbeds.
\label{tab:data}}
\end{table*}

%% file: table/results_cot.tex
\begin{table}[t]
\centering   
\small  
\resizebox{\linewidth}{!}{
\begin{tabular}{l|rrr}
\toprule 
 Method &
 No CoT& {{\shortstack[l]{CoT\\logits before thoughts}}} &{{\shortstack[l]{CoT\\logits after thoughts}}}
\\  
\midrule
First & \diffneg{25}{34}     & \diffneg{28}{35} & \diffneg{24}{35}\\ 
Last & \diffneg{25}{34}      & \diffneg{31}{35} & \diffneg{25}{35}\\ 
Average & \diffneg{29}{34}   & \diffneg{31}{35} & \diffneg{25}{35}\\ 
Sample Avg.& \diffneg{27}{34}& \diffneg{29}{35} & \diffneg{28}{35}\\ 
Sum& \diffneg{32}{34}        & \diffneg{35}{35} & \diffneg{30}{35}
\\ \bottomrule
\end{tabular}}
\caption{Difference between the recall of full decoding and three variants of estimation methods, without or with CoT for ``college mathematics'' in MMLU, using LLaMA 3 Instruct. Methods that underperform or outperform full decoding are highlighted with \colorbox{red!10}{red} or \colorbox{green!10}{green} background.}
\label{tab:cot} 
\spacemagic{\vspace{-1em}}
\end{table}